\documentclass[10pt,twocolumn,letterpaper]{article}

\usepackage{iccv}              

\usepackage{xcolor}
\usepackage{tabularx}
\usepackage{multirow}
\usepackage{float}

\usepackage{appendix}

\usepackage{listings}
\usepackage{colortbl}
\usepackage{arydshln}
\definecolor{codegreen}{rgb}{0,0.6,0}
\definecolor{codegray}{rgb}{0.5,0.5,0.5}
\definecolor{codepurple}{rgb}{0.58,0,0.82}
\definecolor{backcolour}{rgb}{0.95,0.95,0.92}

\lstdefinestyle{mystyle}{
    backgroundcolor=\color{backcolour},   
    commentstyle=\color{codegreen},
    keywordstyle=\color{magenta},
    numberstyle=\tiny\color{codegray},
    stringstyle=\color{codepurple},
    basicstyle=\ttfamily\footnotesize,
    breakatwhitespace=false,         
    breaklines=true,                 
    captionpos=b,                    
    keepspaces=true,                 
    numbers=left,                    
    numbersep=5pt,                  
    showspaces=false,                
    showstringspaces=false,
    showtabs=false,                  
    tabsize=2,
    language=Python,
    morekeywords={as, and, assert, break, class, continue, def, del, elif, else, except, False, finally, for, from, global, if, import, in, is, lambda, None, nonlocal, not, or, pass, raise, return, True, try, while, with, yield},
}

\usepackage{wrapfig}

\definecolor{iccvblue}{rgb}{0.21,0.49,0.74}
\usepackage[pagebackref,breaklinks,colorlinks,allcolors=iccvblue]{hyperref}

\def\paperID{1725} 
\def\confName{ICCV}
\def\confYear{2025}

\title{Rethinking Cross-Modal Interaction in Multimodal Diffusion Transformers}

\author{
    {Zhengyao Lv}$^{1*}$ \quad {Tianlin Pan}$^{2,3*}$ \quad {Chenyang Si}$^{2*}$ \quad {Zhaoxi Chen}$^{4}$ \\
    {Wangmeng Zuo}$^{5}$ \quad {Ziwei Liu}$^{4\dag}$ \quad {Kwan-Yee K. Wong}$^{1\dag}$ \\ \\
    $^1$The University of Hong Kong \quad  $^2$Nanjing University\quad $^3$University of Chinese Academy of Sciences \\ \quad $^4$Nanyang Technological University \quad
    $^5$Harbin Institute of Technology \\
    {\tt\small cszy98@gmail.com} \quad {\tt\small pantianlin23@mails.ucas.ac.cn} \quad {\tt\small chenyang.si@nju.edu.cn} \\  {\tt\small zhaoxi001@ntu.edu.sg} \quad {\tt\small cswmzuo@gmail.com} \quad {\tt\small ziwei.liu@ntu.edu.sg} \quad {\tt\small kykwong@cs.hku.hk}
}

\begin{document}

\twocolumn[{%
    \renewcommand\twocolumn[1][]{#1}%
    \maketitle
    \vspace{-4em}
    \begin{center}
        \centering
        \includegraphics[width=1.0\textwidth]{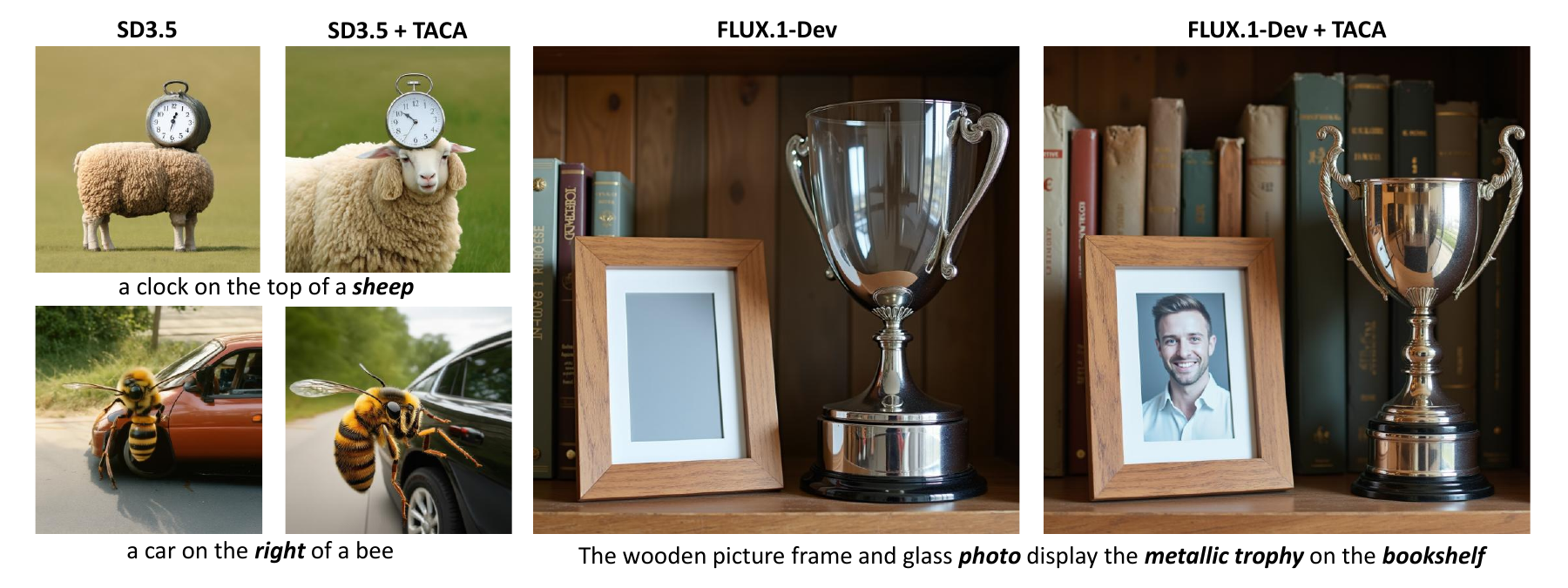}
        \vspace{-2.5em}
        \captionof{figure}{
        We propose \textit{TACA}, a parameter-efficient method that dynamically rebalances cross-modal attention in multimodal diffusion transformers to improve text-image alignment.
        }
        \vspace{-0.5em}
        \label{fig:teaser}
    \end{center}%
    }]
\renewcommand{\thefootnote}{} 
\footnotetext{*Equal Contribution. $^{\dag}$Corresponding Author.}

\begin{abstract}
Multimodal Diffusion Transformers (MM-DiTs) have achieved remarkable progress in text-driven visual generation. However, even state-of-the-art MM-DiT models like FLUX struggle with achieving precise alignment between text prompts and generated content. We identify two key issues in the attention mechanism of MM-DiT, namely 1) the suppression of cross-modal attention due to token imbalance between visual and textual modalities and 2) the lack of timestep-aware attention weighting, which hinder the alignment. To address these issues, we propose \textbf{Temperature-Adjusted Cross-modal Attention (TACA)}, a parameter-efficient method that dynamically rebalances multimodal interactions through temperature scaling and timestep-dependent adjustment. When combined with LoRA fine-tuning, TACA significantly enhances text-image alignment on the T2I-CompBench benchmark with minimal computational overhead. We tested TACA on state-of-the-art models like FLUX and SD3.5, demonstrating its ability to improve text-image alignment in terms of object appearance, attribute binding, and spatial relationships. Our findings highlight the importance of balancing cross-modal attention in improving semantic fidelity in text-to-image diffusion models. Our codes are publicly available at \href{https://github.com/Vchitect/TACA}{https://github.com/Vchitect/TACA}.
\end{abstract}    
\section{Introduction}
Diffusion models~\cite{Ho2020DenoisingDP, Song2020DenoisingDI}, driven by iterative denoising processes, have emerged as a powerful paradigm in generative modeling and various visual generation tasks~\cite{wei2023elite,liu2024smartcontrol,wei2025personalized,zhang2025framepainter}. The field has witnessed significant architectural evolution, beginning with U-Net-based designs~\cite{Ronneberger2015UNetCN} that dominated early diffusion models~\cite{Dhariwal2021DiffusionMB, Rombach2021HighResolutionIS, Podell2023SDXLIL, Nichol2021GLIDETP}. Recent advances introduced transformer-based architectures through Diffusion Transformers (DiT)~\cite{Peebles2022ScalableDM, Chen2023PixArtFT}, demonstrating superior scalability and training stability. This progression culminated in Multimodal Diffusion Transformers (MM-DiT)~\cite{Esser2024ScalingRF}, which unify text and visual tokens through a concatenated self-attention mechanism, resulting in state-of-the-art text-to-image/video models like Stable Diffusion 3/3.5~\cite{Esser2024ScalingRF, StableDiffusion35}, FLUX~\cite{flux2024}, CogVideo~\cite{hong2022cogvideo, yang2024cogvideox}, and HunyuanVideo~\cite{kong2024hunyuanvideo}.

\begin{figure}[h]
    \centering
    \includegraphics[width=0.9\linewidth]{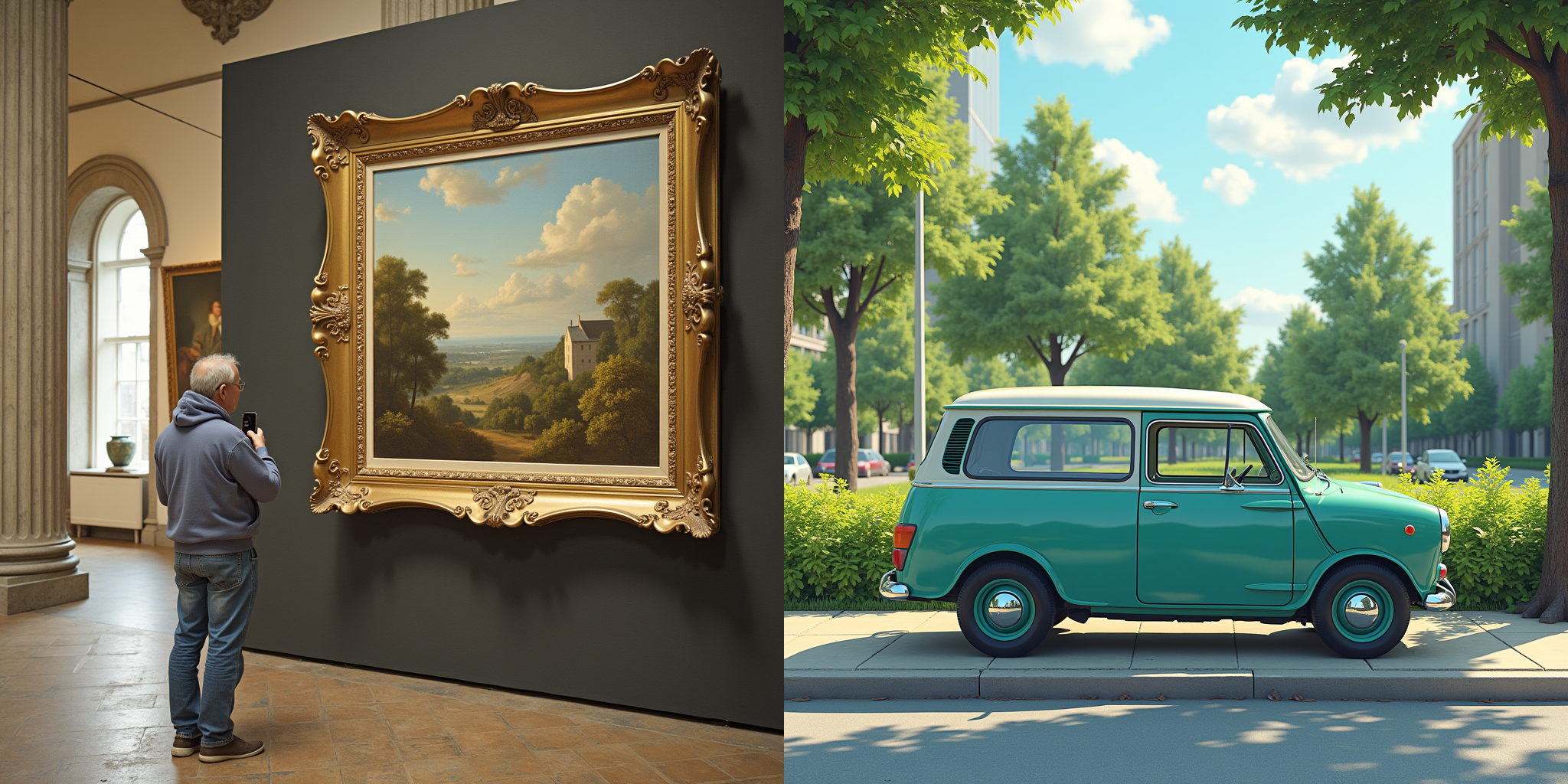}
    \caption{Object missing in text-to-image models. Even in SOTA models like FLUX.1 Dev, we can still observe cases with missing objects. Prompts: ``\textit{The square painting was next to \underline{the round mirror}}", ``\textit{\underline{a blue bench} and a green car}".}
    \label{fig:obj-miss-example}
    \vspace{-1em}
\end{figure}

\begin{figure*}[h]
    \centering
    \begin{minipage}{0.52\textwidth}
        \centering
        \includegraphics[width=1.0\linewidth]{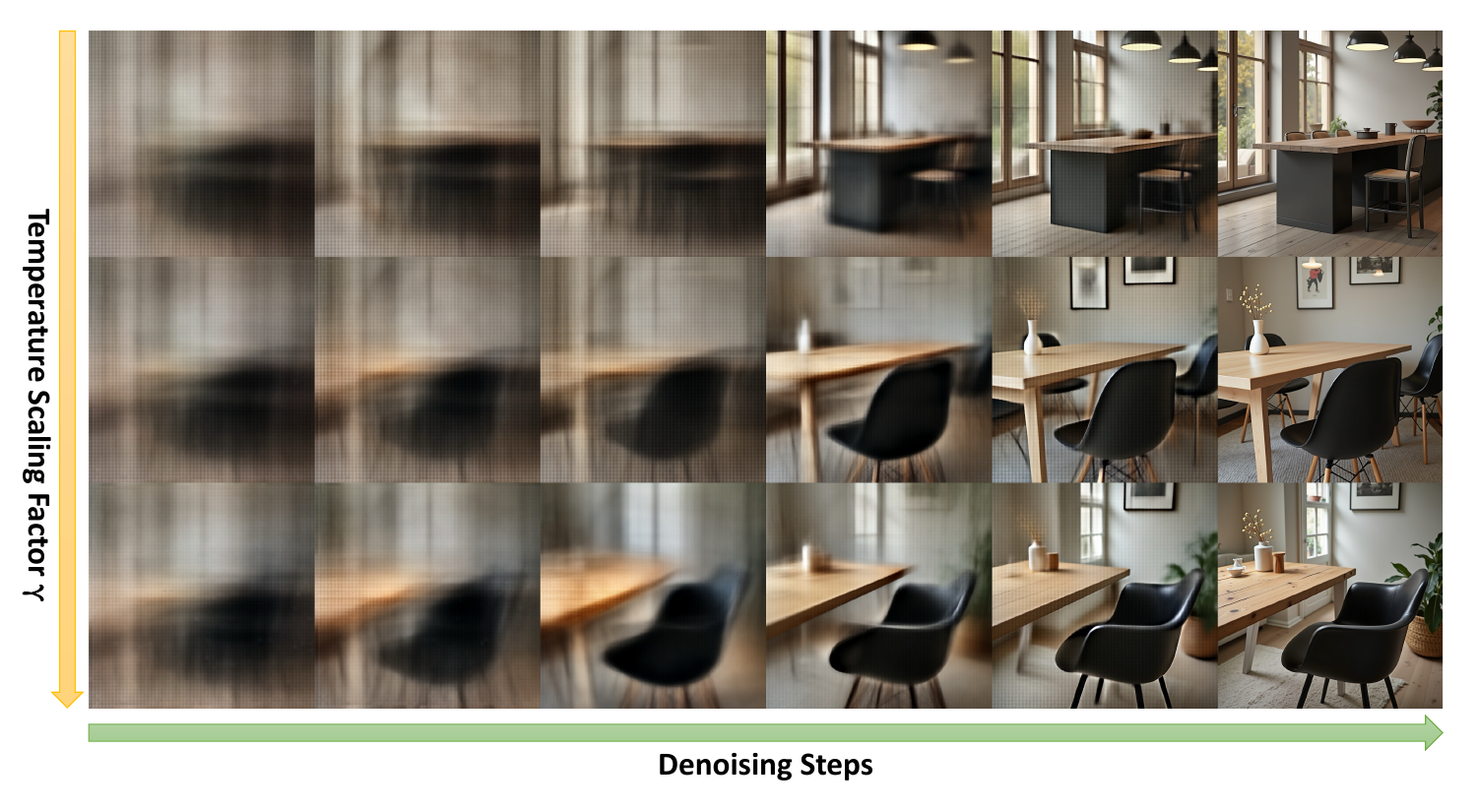}
        \caption{The denoising process. This figure shows the predicted $\boldsymbol{x}_0$ in each step of the denoising process for the prompt ``\textit{The \textbf{black chair} is on the right of the wooden table}'' with FLUX.1 Dev. This observation leads to our hypothesis that visual-text cross-attention plays a more significant role than visual self-attention specifically during these initial steps where the image's overall composition is determined. Additionally, as the temperature scaling factor $\gamma$ increases in the cross-modal section of MM-DiT's unified softmax function, the initial image composition progressively aligns more closely with the corresponding text.}
        \label{fig:denoise_proc}
    \end{minipage}\hfill
    \begin{minipage}{0.45\textwidth}
        \centering
        \includegraphics[width=1.0\linewidth]{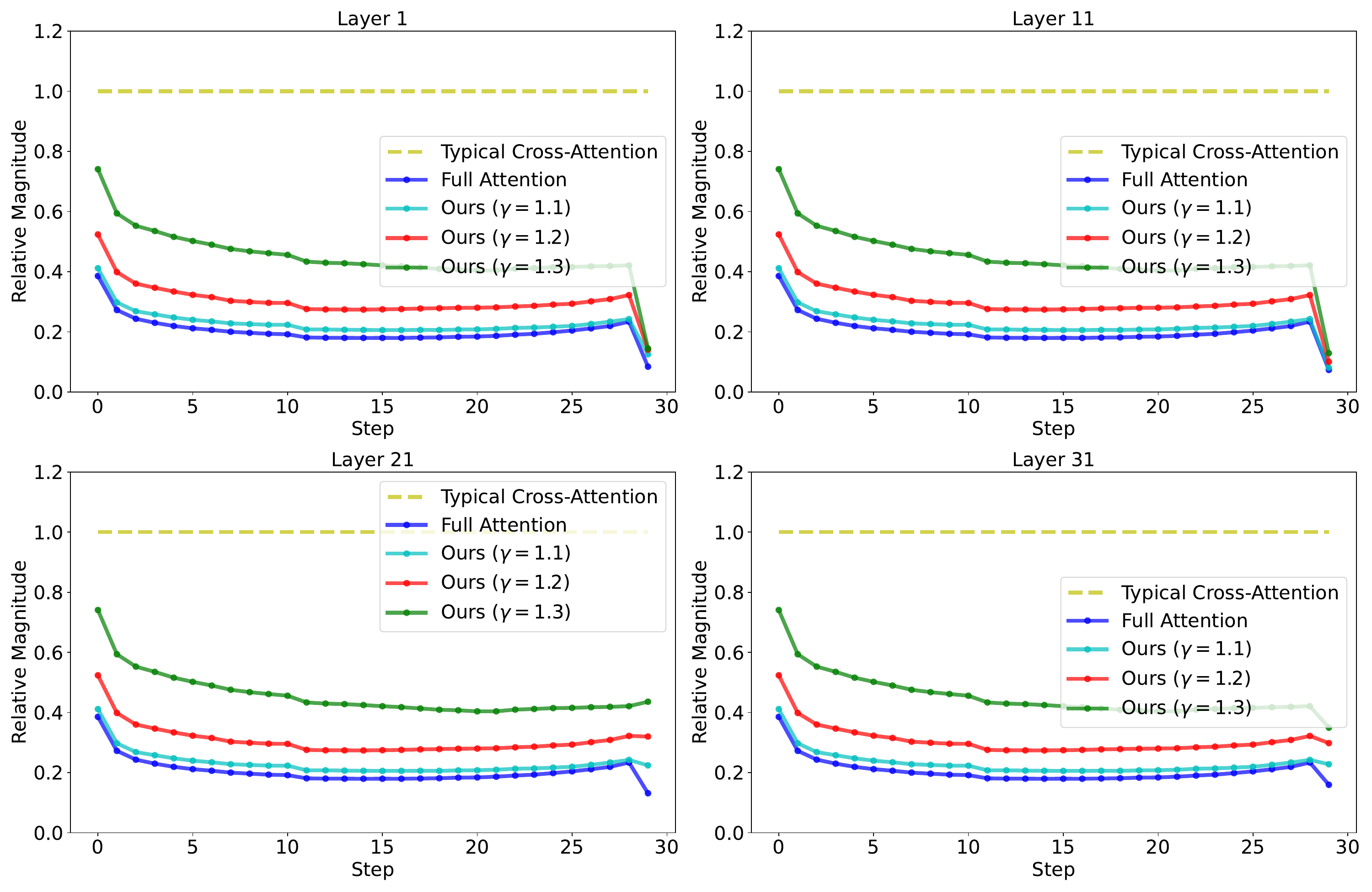}
        \caption{Relative magnitude of visual-text attention between the typical cross attention and MM-DiT full attention (averaged over 50 samples). The numerical asymmetry between the number of  visual and text tokens suppresses the magnitude of cross attention, leading to weak alignment between the generated image and the given text prompt. We can amplify the cross-attention by boosting the coefficient $\gamma$, thereby strengthening the alignment between the image and text.}
        \label{fig:ca-suppress}
    \end{minipage}
\vspace{-1.5em}
\end{figure*}

Although the MM-DiT architecture has undergone significant advancements, state-of-the-art text-to-image models like FLUX still exhibit critical limitations, particularly in generating images with precise semantic alignment (see Fig.~\ref{fig:obj-miss-example}). Analysis of the sampling process reveals that early denoising steps require strong text-visual interaction to create a proper semantic layout, while later steps focus on refining the details. Semantic discrepancies between the text and synthesized images often stem from flawed initial layouts (see Fig.~\ref{fig:denoise_proc}).

In typical U-Net/DiT-based text-to-image diffusion models, the cross-attention block enables modal interaction between textual and visual tokens to synthesize text-aligned images. Our analysis of the attention maps of MM-DiT layers suggests that semantic discrepancies may arise from the suppression of cross-modal attention, specifically due to the numerical asymmetry between the number of visual and text tokens. The overwhelming number of visual tokens can dilute the textual guidance in the unified softmax function of the MM-DiT architecture, resulting in the visual tokens paying significantly less attention to the textual tokens compared to the typical cross-attention paradigm (see Fig.~\ref{fig:ca-suppress}). Furthermore, we noticed that current MM-DiT architectures employ static attention mechanisms with the same weighting for all timesteps, which is ill-suited to the time-varying demands of semantic composition and detail synthesis during the denoising process (see Fig.~\ref{fig:denoise_proc}). This temporal dynamic remains unaddressed in existing approaches, leading to suboptimal modality balancing.

Based on the above observations, we propose \textbf{Temperature-Adjusted Cross-modal Attention} \textbf{(TACA)}, a straightforward yet effective enhancement to the MM-DiT attention mechanism. Our approach introduces two key innovations, namely \textbf{(1)} modality-specific temperature scaling to mitigate cross-attention suppression, and \textbf{(2)} timestep-dependent adjustments to cross-modal interactions. TACA only requires a temperature coefficient $\gamma(t)$ that adapts to the denoising timesteps, allowing for easy implementation with minimal code modifications. To mitigate potential artifacts introduced by amplified cross-attention, we complement TACA with Low-Rank Adaptation (LoRA)~\cite{Hu2021LoRALA} fine-tuning for distributional alignment, helping the model generate images that better match real-world distributions.

Experiments on T2I-CompBench~\cite{huang2023t2icompbench} validate the effectiveness of our method across various model architectures. For FLUX.1-Dev, incorporating TACA results in substantial improvements, yielding relative gains of 16.4\% in spatial relationship understanding and 5.9\% in shape accuracy. Similarly, when applied to SD3.5-Medium, TACA boosts spatial relationship accuracy by 28.3\% and shape accuracy by 2.9\%. These benchmark results, combined with the visual improvements shown in Fig.~\ref{fig:teaser}, highlight a significant enhancement in text-image alignment achieved by our approach.

In summary, our principal contributions are:
\begin{itemize}
    \item We systematically analyze MM-DiT's unified attention mechanism, and reveal cross-attention suppression and timestep insensitivity being two key factors limiting text-image alignment in text-to-image generation.
    \item We propose TACA, the first approach to dynamically balance multimodal interactions through temperature scaling and temporal adaptation in diffusion transformers.
    \item Extensive benchmark results demonstrate that TACA can effectively improve semantic alignment with minimal computational overhead.
\end{itemize}
\section{Related Work}
\subsection{Diffusion Transformers}
A central challenge in developing transformer-based text-to-image/video (T2I, T2V) diffusion models lies in the effective integration of multimodal data, primarily text and visual information.  Several approaches, including Diffusion Transformers (DiT~\cite{Peebles2022ScalableDM}), CrossDiT (PixArt-$\alpha$~\cite{Chen2023PixArtFT}), and MM-DiT (Stable Diffusion 3~\cite{Esser2024ScalingRF}), tackle this challenge with distinct methods for cross-modal interaction and text-image alignment.

\noindent\textbf{The original DiT}~\cite{Peebles2022ScalableDM} introduced transformers~\cite{Vaswani2017AttentionIA, Dosovitskiy2020AnII} as replacements for U-Net backbones~\cite{Ronneberger2015UNetCN} in diffusion models~\cite{Ho2020DenoisingDP, Song2020DenoisingDI}. While not inherently multimodal, DiT established critical conditioning mechanisms via adaptive layer normalization (adaLN)~\cite{Perez2017FiLMVR}. This technique modulates transformer activations using timestep embeddings and class labels, enabling controlled generation based on single-modality inputs. While effective for class-conditional generation, DiT lacks explicit mechanisms for text-image alignment, limiting its applicability in multimodal tasks.

\noindent\textbf{CrossDiT (PixArt-$\alpha$})~\cite{Chen2023PixArtFT} introduced cross-modal fusion by integrating text-guided cross-attention into the DiT backbone. In this framework, cross-attention replaces adaLN for text conditioning, which enables dynamic per-token modulation based on linguistic semantics. However, CrossDiT uses a unidirectional update approach that prevents the image from influencing the textual representation. This hinders its ability to model feedback loops and nuanced interdependencies between the text and generated image.

\noindent\textbf{MM-DiT (Stable Diffusion 3)}~\cite{Esser2024ScalingRF} represents a paradigm shift by introducing bidirectional cross-modal attention and a unified token space for text and visual modalities. By concatenating text and image tokens into a single sequence and employing a decomposed attention matrix, MM-DiT enables full self-attention across modalities, capturing complex inter-modal relationships. Besides, the integration of multiple text encoders (e.g., CLIP~\cite{Radford2021LearningTV} and T5~\cite{Raffel2019ExploringTL}) further improves the model's ability to understand and generate text with greater accuracy.

\subsection{Text-to-Image Alignment}
Prior research has explored generating images from text prompts using pre-trained models without requiring further training. Some employ techniques such as CLIP-guided optimization~\cite{liu2021fusedream, galatolo2021clipglass} to align images with text by optimizing CLIP scores within the model's latent space. Additionally, cross-attention-based approaches~\cite{chen2023trainingfree} are used to enhance spatial layout and details in generated images, thereby improve adherence to the textual description's structure.

Additionally, more recent research has explored augmenting guidance-based models to enhance semantic control, primarily through layout planning modules~\cite{Xie2023BoxDiffTS, Li2023GLIGENOG, Dahary2024BeYB, Chen2023TrainingFreeLC, Phung2023GroundedTS, Kim2023DenseTG, Yang2024MasteringTD} and feedback-driven optimization~\cite{Sun2023DreamSyncAT, Fan2023DPOKRL, Black2023TrainingDM}.  Another direction involves attention-based methods~\cite{Chefer2023AttendandExciteAS, Rassin2023LinguisticBI, Meral2023CONFORMCI, Wang2023TokenComposeTD, Agarwal2023ASTARTA, Li2023DivideB} that modify or constrain the attention maps within U-Net models to improve textual alignment. However, these attention-based techniques generally do not readily translate to contemporary DiT-based architectures.
\section{Methodology}
\subsection{Preliminaries}
\noindent\textbf{Diffusion-based generative models} operate through a forward diffusion process and a reverse denoising process~\cite{Ho2020DenoisingDP}. The forward process systematically degrades data samples through gradual noise injection, while the reverse process learns to recover the original data structure through iterative refinement.

The diffusion mechanism progressively corrupts training samples $\boldsymbol{x}_0 \sim q(\boldsymbol{x}_0)$ over $T$ discrete timesteps according to a predetermined variance schedule $\{\beta_t\}_{t=1}^T$. This corruption follows a Markov chain where each transition adds isotropic Gaussian noise:
\begin{align}
    q(\boldsymbol{x}_t|\boldsymbol{x}_{t-1}) = \mathcal{N}\left(\boldsymbol{x}_t; \sqrt{1-\beta_t}\boldsymbol{x}_{t-1}, \beta_t\mathbf{I}\right).
\end{align}

The denoising phase constitutes a learned reversal of this progressive corruption. This reverse process estimates the ancestral distribution $q(\boldsymbol{x}_{t-1}|\boldsymbol{x}_t)$ by learning:
\begin{align}
    p_\theta(\boldsymbol{x}_{t-1}|\boldsymbol{x}_t) = \mathcal{N}\left(\boldsymbol{x}_{t-1}; \boldsymbol{\mu}_\theta(\boldsymbol{x}_t, t), \sigma_t^2\mathbf{I}\right),
\end{align}
where $\sigma_t^2$ is typically fixed as $\beta_t$ or $\tilde{\beta}_t = \frac{1 - \bar{\alpha}_{t-1}}{1 - \bar{\alpha}_t}\beta_t$ with $\bar{\alpha}_t = \prod_{s=1}^t (1-\beta_s)$. The mean $\boldsymbol{\mu}_\theta$ is derived through a noise prediction network $\epsilon_\theta$. This network, conventionally implemented as a time-conditional U-Net~\cite{Ronneberger2015UNetCN} or vision transformers~\cite{Peebles2022ScalableDM, Chen2023PixArtFT, Esser2024ScalingRF} in more recent works, is optimized to predict the noise component presents in $\boldsymbol{x}_t$, enabling precise incremental denoising.

\noindent\textbf{Multimodal Diffusion Transformer (MM-DiT)}~\cite{Esser2024ScalingRF} is a novel approach to adopt transformers as the noise prediction network in diffusion models. The MM-DiT architecture concatenates text and visual tokens into a single input sequence after projecting both modalities to a shared dimensional space. The concatenated sequence undergoes multi-head self-attention where every token attends to all others, regardless of modality. Mathematically, if we use $H$ to denote the number of attention heads, $N_x$ and $N_c$ to denote the sequence length of visual and text tokens respectively, and $D$ to denote the dimension of the token embeddings, then for visual tokens $\boldsymbol{x} \in \mathbb{R}^{H \times N_x \times D}$ and text tokens $\boldsymbol{c} \in \mathbb{R}^{H \times N_c \times D}$, we have:
\begin{equation}
\boldsymbol Q = \begin{pmatrix}
\boldsymbol W^Q_c\boldsymbol c \\
\boldsymbol W^Q_x\boldsymbol x 
\end{pmatrix},
\ 
\boldsymbol K = \begin{pmatrix}
\boldsymbol W^K_c\boldsymbol c \\ 
\boldsymbol W^K_x\boldsymbol x
\end{pmatrix},
\ 
\boldsymbol V = \begin{pmatrix}
\boldsymbol W^V_c\boldsymbol c \\
\boldsymbol W^V_x\boldsymbol x 
\end{pmatrix},
\end{equation}
and
\begin{equation}
\text{Attention}(\boldsymbol Q,\boldsymbol K,\boldsymbol V) = \text{softmax}\left(\frac{\boldsymbol Q \boldsymbol K^T}{\sqrt{D}}\right)\boldsymbol V,
\label{eq:scale-dot-prod-attn}
\end{equation}
where the $\boldsymbol Q\boldsymbol K^T$ term can be expanded to 
\begin{align}
\boldsymbol Q\boldsymbol K^T &= \begin{pmatrix}
\boldsymbol W^Q_c\boldsymbol c(\boldsymbol  W^K_{c}\boldsymbol c)^T & \boldsymbol W^Q_c\boldsymbol c(\boldsymbol W^K_x\boldsymbol x)^T \\
\boldsymbol W^Q_x\boldsymbol x(\boldsymbol W^K_c\boldsymbol c)^T & \boldsymbol W^Q_x\boldsymbol x(\boldsymbol W^K_x\boldsymbol x)^T
\end{pmatrix}\\
&= \begin{pmatrix}
\boldsymbol Q_{\mathrm{txt}}\boldsymbol K^T_{\mathrm{txt}} & \boldsymbol Q_{\mathrm{txt}}\boldsymbol K^T_{\mathrm{vis}} \\
\boldsymbol Q_{\mathrm{vis}}\boldsymbol K^T_{\mathrm{txt}} & \boldsymbol Q_{\mathrm{vis}}\boldsymbol K^T_{\mathrm{vis}} 
\end{pmatrix}.
\label{eq:qkt-mat}
\end{align}
As we can see in Eq \ref{eq:qkt-mat}, this MM-DiT formulation allows four interaction types: text-text, text-visual, visual-text, and visual-visual attentions, all within a single operation. 

\subsection{Suppression of Cross-Attention and Timestep-Insensitive Weighting in MM-DiT} 

\label{sec:drawbacks}
While the unified attention mechanism of MM-DiT provides computational efficiency through joint modality processing, it introduces inherent issues when balancing different modalities. 

\noindent\textbf{Suppression of Cross-Attention\quad}This issue stems from the numerical asymmetry between the number of visual and text tokens ($N_{x} \gg N_{c}$), which creates a systematic bias in attention weight distribution. 
Consider the attention computation for visual tokens in Eq.~\ref{eq:qkt-mat}. Each visual token's attention weights over text tokens ($\boldsymbol Q_{\mathrm{vis}}\boldsymbol K_{\mathrm{txt}}^T$) must compete against visual-visual interactions ($\boldsymbol Q_{\mathrm{vis}}\boldsymbol K_{\mathrm{vis}}^T$) in the softmax denominator. Formally, the probability of the $i$-th visual token attending to the $j$-th text token guidance becomes:
\begin{equation}
P_{\mathrm{vis-txt}}^{(i,\,j)} = \frac{ e^{s_{ij}^{\mathrm{vt}}/\tau}}{\sum_{k=1}^{N_{\mathrm{txt}}} e^{s_{ik}^{\mathrm{vt}}/\tau} + \sum_{k=1}^{N_{\mathrm{vis}}} e^{s_{ik}^{\mathrm{vv}}/\tau}},
\label{eq:unified-softmax}
\end{equation}
where $s_{ik}^{\mathrm{vt}}$ $=$ $\boldsymbol Q^{(i)}_{\mathrm{vis}}\boldsymbol K_{\mathrm{txt}}^{T\,(k)}/\sqrt{D}$ and $s_{ik}^{\mathrm{vv}}$ $=$ $\boldsymbol Q^{(i)}_{\mathrm{vis}}\boldsymbol K_{\mathrm{vis}}^{T\,(k)}/\sqrt{D}$. When $N_{\mathrm{vis}} \gg N_{\mathrm{txt}}$, the sum over visual-visual interactions dominates the denominator, even if individual $s_{ik}^{\mathrm{vv}}$ values are modest. For example, when using FLUX.1 Dev~\cite{flux2024} to generate a 1024 $\times$ 1024 image, we have $N_{\mathrm{vis}} / N_{\mathrm{txt}} = 4096 / 512 = 8$. In this case, the visual-text cross-attention probabilities would be much lower than in the typical paradigm:
\begin{align}
P_{\mathrm{vis-txt}}^{(i,\,j)} &\approx \frac{ e^{s_{ij}^{\mathrm{vt}}/\tau}}{\sum_{k=1}^{N_{\mathrm{vis}}} e^{s_{ik}^{\mathrm{vv}}/\tau}} \ \hfill (\text{Full Attention}) 
\label{eq:full-attn} \\
&\ll \frac{ e^{s_{ij}^{\mathrm{vt}}/\tau}}{\sum_{k=1}^{N_{\mathrm{txt}}} e^{s_{ik}^{\mathrm{vt}}/\tau}} \ \hfill (\text{Typical Cross-Attention})
\label{eq:ca-suppress}
\end{align}
This suppression of $P_{\mathrm{vis-txt}}$, which can be observed in Fig.~\ref{fig:ca-suppress}, weakens the alignment between visual and textual features. The model struggles to effectively leverage textual guidance to refine visual representations because the influence of the text tokens is diluted by the overwhelming presence of visual tokens.  Crucial semantic relationships encoded in the text may be overlooked, leading to a visual representation that is less informed by the corresponding textual description, like the bad cases shown in Fig.~\ref{fig:obj-miss-example}.

\noindent\textbf{Timestep-Insensitive QK Weighting\quad}MM-DiT's current architecture employs timestep-agnostic projection of latent states into query and key vectors. This approach fails to account for the evolving influence of textual guidance throughout the denoising process. As illustrated in Fig.~\ref{fig:denoise_proc}, the initial denoising steps are crucial for establishing the image's global layout, heavily influenced by the text prompt. Consequently, the cross-attention mechanism, responsible for integrating textual information, should be weighted more heavily than visual self-attention during these early stages. MM-DiT's static weighting strategy, therefore, limits its ability in optimally leveraging textual guidance and adapting to the changing demands of the denoising process.

Formally, when $t$ is large (i.e., early in the denoising process) and cross-modal guidance should dominate, $s_{ik}^{\mathrm{vt}}$ values fail to receive proportionally larger magnitudes compared to $s_{ik}^{\mathrm{vv}}$. Since $\boldsymbol W^Q$ and $\boldsymbol W^K$ are optimized for global performance across all timesteps, they cannot focus on amplifying visual-text interactions in the early stages. This potentially leads to the overall layout of the generated image not aligning with the text prompt.

\begin{figure}[h]
    \centering
    \includegraphics[width=1.0\linewidth]{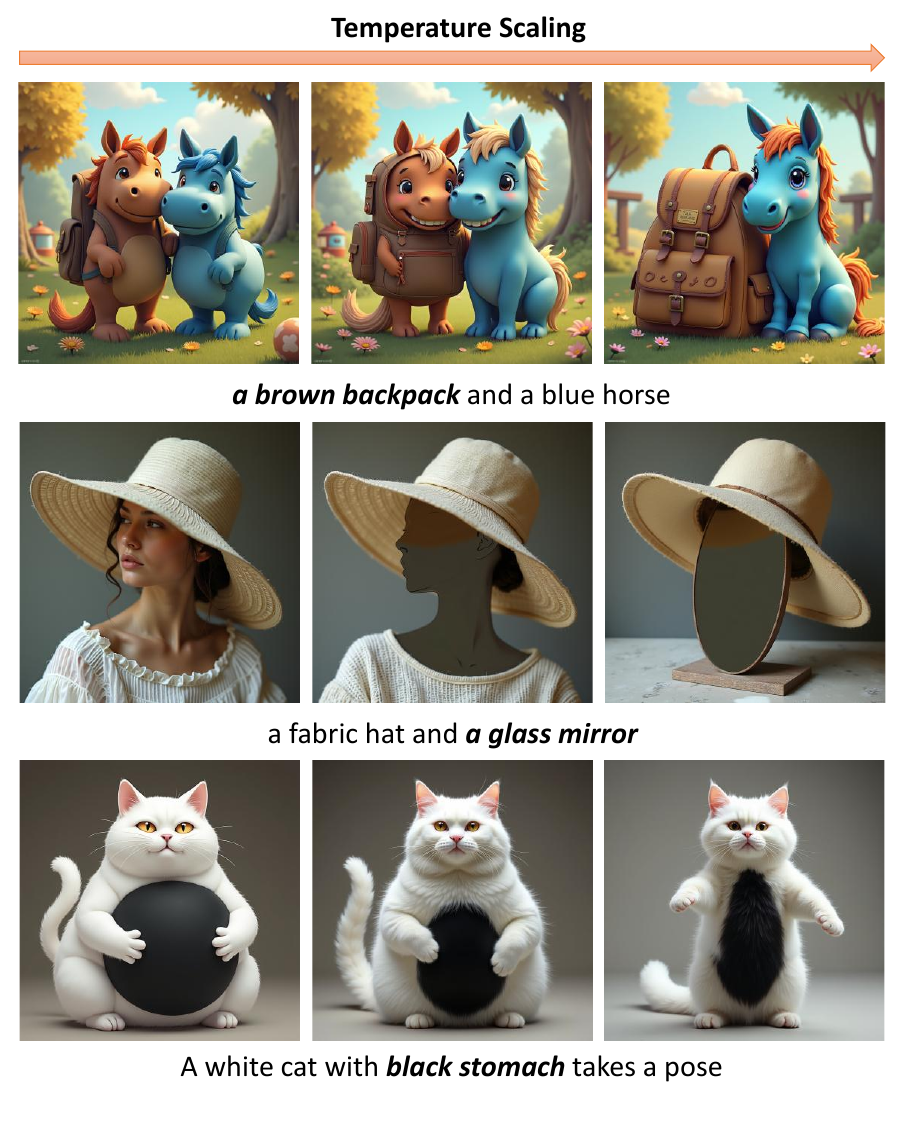}
    \vspace{-3em}
    \caption{Temperature scaling helps text-image alignment. From this figure, we can see that as the temperature scaling factor $\gamma$ increases, the characteristics of ``\textit{brown backpack}'', ``\textit{glass mirror}'' and ``\textit{black stomach}'' become more obvious.}
    \label{fig:various_gamma}
    \vspace{-1.5em}
\end{figure}

\subsection{Temperature-Adjusted Cross-modal Attention}

To address the issues mentioned in Section \ref{sec:drawbacks}, we propose \textbf{Temperature-Adjusted Cross-modal Attention (TACA)}, a simple yet effective modification to the attention mechanism of MM-DiT. Our approach introduces two key innovations, namely \textit{modality-specific temperature scaling} and \textit{timestep-dependent adjustment of cross-modal interactions}.

\noindent\textbf{Modality-Specific Temperature Scaling\quad}To mitigate the suppression of cross-attention caused by the dominance of visual tokens ($N_{\mathrm{vis}} \gg N_{\mathrm{txt}}$), we amplify the logits of visual-text interactions through a \textit{temperature coefficient} $\gamma > 1$. The modified attention probability for visual-text interaction becomes:
\begin{equation}
P_{\mathrm{vis-txt}}^{(i,\,j)} = \frac{ e^{{\color{blue}\gamma} s_{ij}^{\mathrm{vt}}/\tau}}{\sum_{k=1}^{N_{\mathrm{txt}}} e^{{\color{blue}\gamma} s_{ik}^{\mathrm{vt}}/\tau} + \sum_{k=1}^{N_{\mathrm{vis}}} e^{s_{ik}^{\mathrm{vv}}/\tau}},
\end{equation}

This scaling effectively rebalances the competition in softmax by increasing the relative weights of cross-modal interactions. The $\gamma$ coefficient acts as a \textit{signal booster} for text-guided attention. As shown in Fig.~\ref{fig:various_gamma}, the generated image and text prompt become more consistent as $\gamma$ increases. 

\begin{figure}
    \centering
    \includegraphics[width=1.0\linewidth]{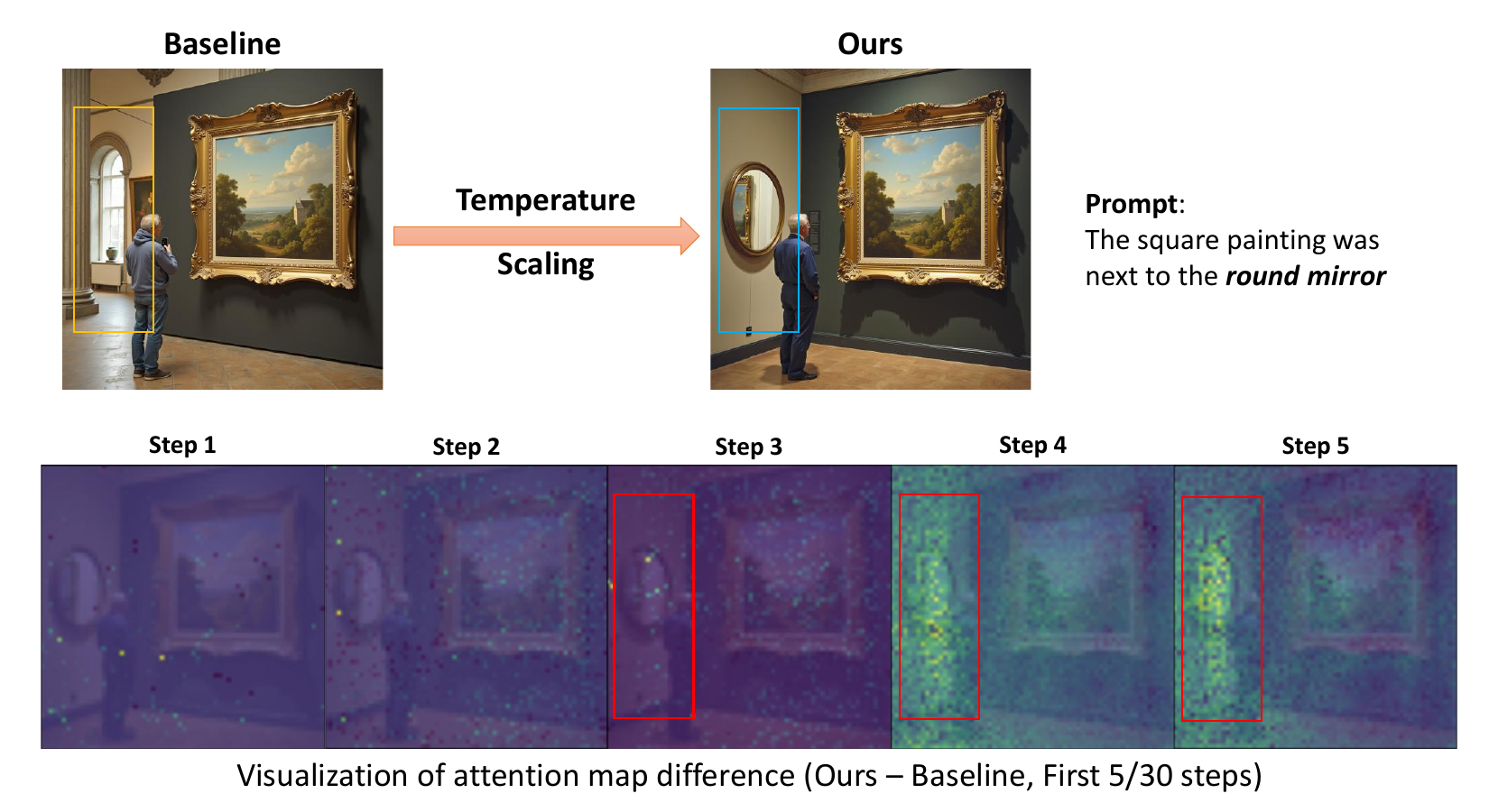}
    \vspace{-2em}
    \caption{Attention map differences. We conducted a visualization of the alterations in the visual-text attention map during the initial stages of the denoising process, as influenced by our proposed method. In contrast to the baseline, our approach substantially amplifies the attention directed toward the text in the early steps.}
    \label{fig:attn-map}
    \vspace{-2em}
\end{figure}

\noindent\textbf{Timestep-Dependent Adjustment\quad}To compensate for the insensitivity of QK weights with respect to the timestep, we make $\gamma$ timestep-dependent to account for the varying importance of cross-attention during denoising based on the observations in Fig.~\ref{fig:denoise_proc}. Specifically, we employ a piecewise function:
\begin{equation}
\gamma(t) = \begin{cases}
\gamma_0 & t \geq t_{\mathrm{thresh}} \\
1 & t < t_{\mathrm{thresh}}
\end{cases}
\end{equation}
where $t_{\mathrm{thresh}}$ is a threshold for the timestep that separates the \textit{layout formation} and \textit{detail refinement} phases. This design aligns with the denoising dynamics where early steps (i.e., large $t$) require strong text guidance to establish image composition and later steps (i.e., small $t$) focus on visual details when self-attention dominates. By effecting the attention map, image tokens can better attend to the relevant text tokens, as shown in Fig.~\ref{fig:attn-map}.



Notably, TACA introduces no new learnable parameters, with the temperature scaling implemented via a simple element-wise operation during attention computation. The $\gamma_0$ and $t_{\mathrm{thresh}}$ parameters can be tuned through minimal ablation studies, making our approach both efficient and practical for deployment in existing MM-DiT architectures.

\noindent\textbf{LoRA Training for Artifact Suppression\quad}While temperature scaling in TACA significantly improves text-image alignment, the amplified cross-modal attention logits can alter the output distribution of the denoising process, occasionally introducing artifacts such as distorted object boundaries or inconsistent textures. To mitigate this, we employ Low-Rank Adaptation (LoRA)~\cite{Hu2021LoRALA} to fine-tune the model, encouraging it to recover the real image distribution while preserving the benefits of temperature scaling.

We apply LoRA to the attention layers of MM-DiT, where the temperature scaling exerts the most direct influence. For a weight matrix $\boldsymbol W \in \mathbb{R}^{d \times k}$, LoRA adaptation is formulated as
\begin{equation}
\boldsymbol W' = \boldsymbol W + \alpha \cdot \boldsymbol B \boldsymbol A, \quad \boldsymbol B \in \mathbb{R}^{d \times r}, \ \boldsymbol A \in \mathbb{R}^{r \times k}
\end{equation}
where $r \ll \min(d,k)$ is the rank of the adaptation, and $\alpha$ scales the low-rank update. Only $\boldsymbol{B}$ and $\boldsymbol{A}$ are trainable during fine-tuning, keeping the original $\boldsymbol{W}$ frozen.
\begin{figure*}[h]
    \centering
    \begin{subfigure}[b]{\linewidth}
        \includegraphics[width=.93\linewidth]{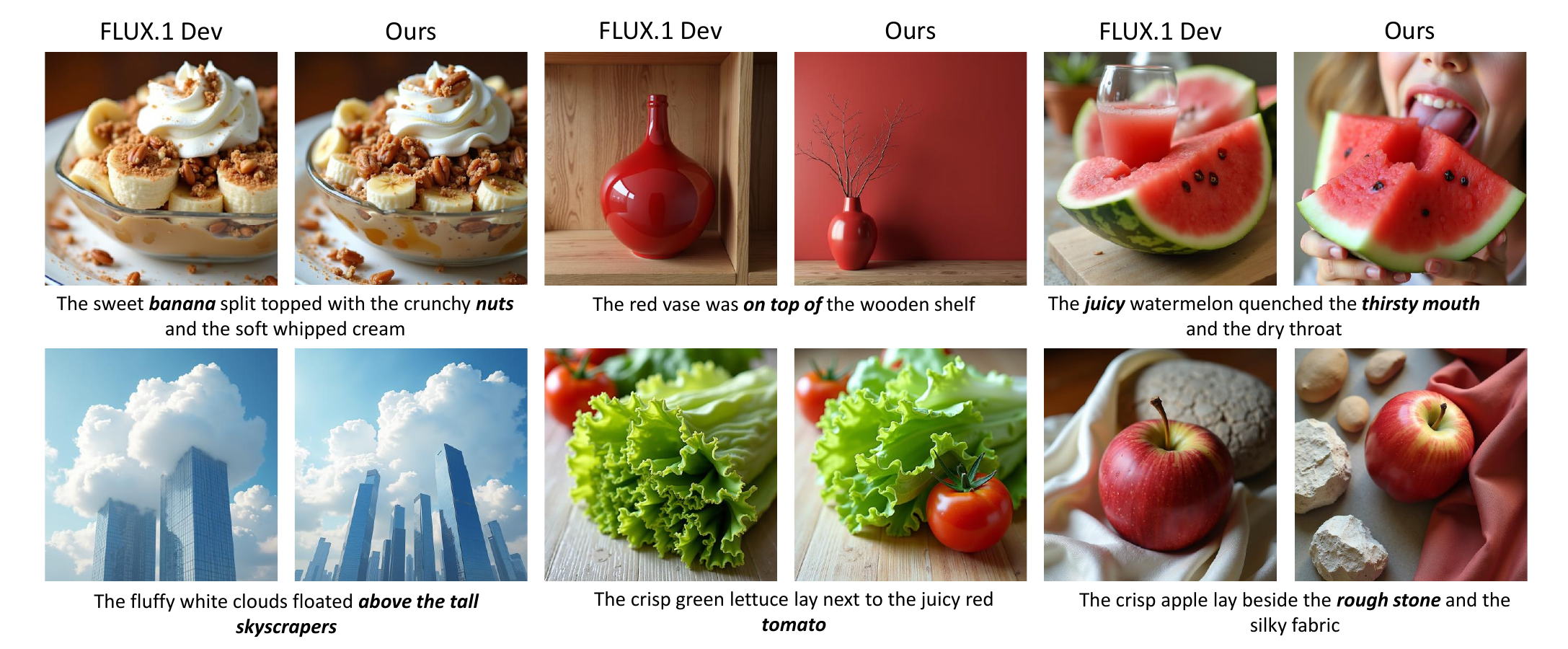}
        \label{fig:flux-examples}
    \end{subfigure}
    \begin{subfigure}[b]{\linewidth}
        \includegraphics[width=.93\linewidth]{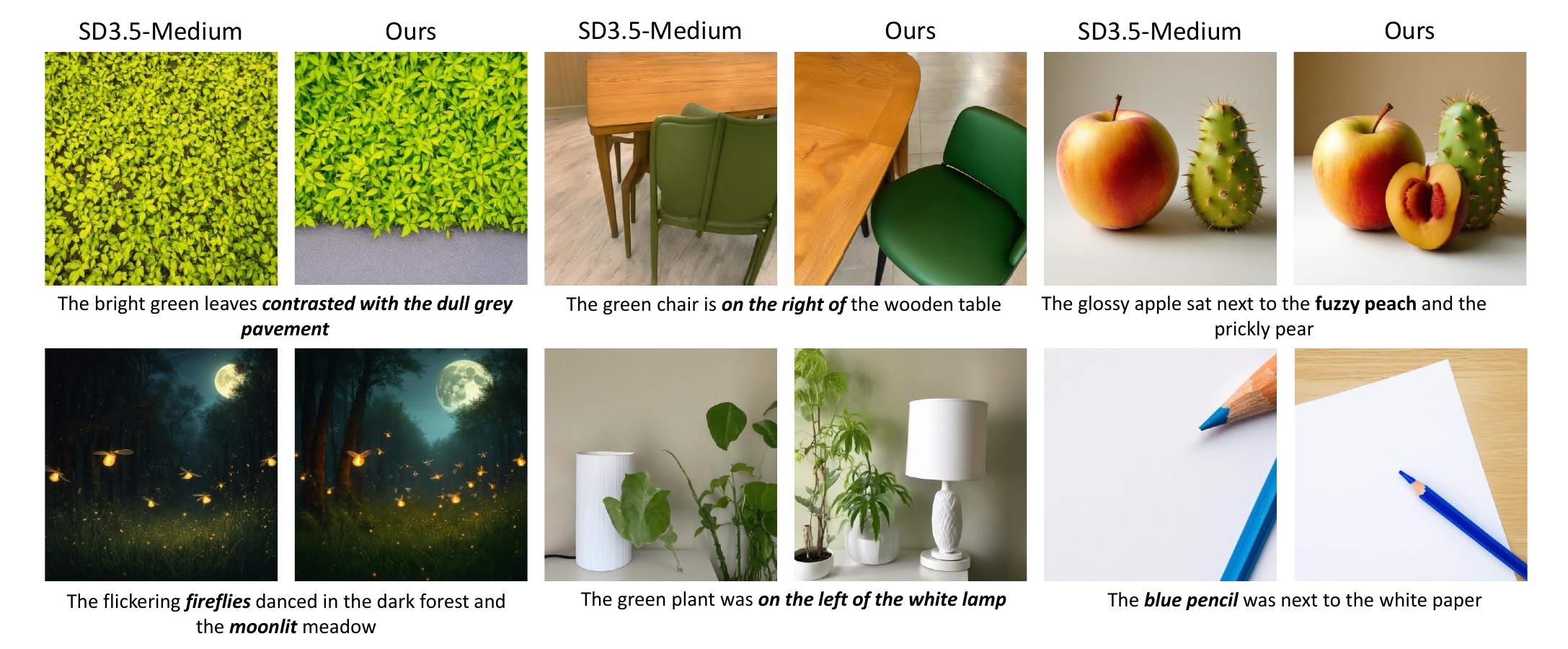}
        \label{fig:sd-examples}
    \end{subfigure}
    \vspace{-2.5em}
    \caption{Comparison of samples generated by FLUX.1 Dev and Stable Diffusion 3.5 Medium with and without TACA. 
    }
    \vspace{-0.6em}
    \label{fig:combined-examples}
\end{figure*}

\begin{table*}[ht]
\caption{Comparison of alignment evaluation on T2I-CompBench~\cite{huang2023t2icompbench} for FLUX.1-Dev-based and SD3.5-Medium-based models. The best results for each model group are highlighted in \textbf{bold}.}
\vspace{-0.75em}
\label{tab:main-results}
\centering
\begin{tabular}{ccccccc}
\toprule
\multirow{2}{*}{Model}           & \multicolumn{3}{c}{Attribute Binding} & \multicolumn{2}{c}{Object Relationship} & \multirow{2}{*}{Complex $\uparrow$} \\ \cline{2-6}
                                 & Color $\uparrow$      & Shape $\uparrow$      & Texture $\uparrow$     & Spatial $\uparrow$          & Non-Spatial $\uparrow$          &                          \\ \hline
\multicolumn{1}{c|}{FLUX.1-Dev} & 0.7678     & 0.5064     & 0.6756      & 0.2066           & 0.3035               & 0.4359                   \\
\multicolumn{1}{c|}{FLUX.1-Dev + TACA ($r = 64$)} & \textbf{0.7843}     & \textbf{0.5362}     & \textbf{0.6872}      & \textbf{0.2405}           & 0.3041               & \textbf{0.4494}                   \\
\multicolumn{1}{c|}{FLUX.1-Dev + TACA ($r = 16$)}  & 0.7842     & 0.5347     & 0.6814      & 0.2321           & \textbf{0.3046}               & 0.4479                   \\ \hline
\multicolumn{1}{c|}{SD3.5-Medium} & 0.7890     & 0.5770     & 0.7328      & 0.2087           & 0.3104               & 0.4441                   \\
\multicolumn{1}{c|}{SD3.5-Medium + TACA ($r = 64$)} & \textbf{0.8074}     & \textbf{0.5938}     & \textbf{0.7522}      & \textbf{0.2678}           & 0.3106               & 0.4470                   \\
\multicolumn{1}{c|}{SD3.5-Medium + TACA ($r = 16$)} & 0.7984     & 0.5834     & 0.7467      & 0.2374           & \textbf{0.3111}               & \textbf{0.4505}                   \\ \bottomrule
\end{tabular}
\vspace{-1.5em}
\end{table*}
\section{Experiments}
\vspace{-0.5em}
\subsection{Experiment Settings}
\vspace{-0.5em}
\noindent\textbf{Evaluation Metrics and Datasets\quad}We evaluate our method on the T2I-CompBench benchmark~\cite{huang2023t2icompbench}, a comprehensive evaluation suite for text-to-image alignment. All experiments use the LAION dataset~\cite{Schuhmann2021LAION400MOD} with captions refined by the LLaVA model~\cite{Liu2023VisualIT} to enhance semantic precision. We randomly sampled 10K image-text pairs as the training dataset for our LoRA model. To ensure reproducibility throughout all evaluation phases, the random seed is fixed to 42, while all other parameters remain at their default values as provided by the Diffusers library~\cite{von-platen-etal-2022-diffusers}.

\noindent\textbf{Implementation Details\quad}We conduct experiments on a single NVIDIA A100 80GB GPU using the ai-toolkit codebase~\cite{AIToolkit}, with LoRA adapters implemented for FLUX.1 Dev~\cite{flux2024} and SD3.5 Medium~\cite{StableDiffusion35} models. We adopt the AdamW optimizer with a learning rate of $1 \times 10^{-4}$ and a batch size of $4$ for training. We evaluate two LoRA configurations: $(r, \alpha)$ = $(16, 16)$ and $(64, 64)$.

To emphasize semantic alignment, we sample timesteps $t \geq t_{\mathrm{thresh}} = 970$ within the range $t \in (0, 1000)$. In the flow matching scheduler, a $30$-step denoising process allocates the first three steps to $t \in (970, 1000]$ (i.e., the initial $10\%$ of the diffusion process), while the remaining $27$ steps cover $t \in [0, 970)$. Setting $t_{\mathrm{thresh}} = 970$ focuses training on these early steps where semantic information is most prominent.

Under the flow-matching paradigm~\cite{Lipman2022FlowMF}, the model predicts velocity $\boldsymbol v$ instead of noise $\epsilon$. We fine-tune it with the following velocity prediction loss:  
\begin{equation}
\small
    \mathcal{L} = \mathbb{E}_{\boldsymbol{x}_0, \ t\geq t_{\mathrm{thresh}}} \left[ \| \boldsymbol v(\boldsymbol x_t, t ) - \boldsymbol v_\theta(\boldsymbol{x}_t, t, \mathcal{P}_{\text{txt}}, \gamma(t)) \|_2^2 \right],
\end{equation}
where $\mathcal{P}_{\text{txt}}$ represents text prompts and $\gamma(t)$ induces the modified temperature coefficient. For benchmark results, we set the base temperature scaling factor as $\gamma_0 = 1.2$, which is selected in Section~\ref{sec:ablation-study}. This formulation ensures the model learns the correct velocity field while adapting to temperature-scaled attention.

\subsection{Main Results}
\noindent\textbf{Quantitative Comparison\quad}
To quantitatively evaluate the effectiveness of our proposed TACA, we conduct a comprehensive comparison against baseline models. Table \ref{tab:main-results} presents the alignment performance of FLUX.1-Dev and SD3.5-Medium models, respectively, with and without the integration of TACA.  For FLUX.1-Dev, the incorporation of TACA, particularly with a rank $r=64$, consistently improves performance across all Attribute Binding metrics and Spatial Relationship. Similarly, for SD3.5-Medium, TACA with $r=64$ yields significant gains in Attribute Binding and Spatial Relationship, and TACA with $r=16$ achieves the best performance on Non-Spatial Relationship and Complex prompt evaluation. These results demonstrate that TACA effectively enhances the alignment capabilities of different MM-DiT models across various dimensions of text-to-image generation quality.

\noindent\textbf{Image Quality Evaluation\quad}We use widely adopted image quality assessment models MUSIQ~\cite{ke2021musiqmultiscaleimagequality} and MANIQA~\cite{yang2022maniqamultidimensionattentionnetwork} to evaluate visual quality. As shown in the Table~\ref{tab:iqa}, TACA improves text-image alignment without sacrificing image quality on both SD3.5 and FLUX. Additionally, Fig.~\ref{fig:combined-examples} presents further visual comparison results.

\begin{table}[h]
\centering
\vspace{-0.75em}
\caption{Results of image quality assessment.}
\vspace{-0.75em}
\label{tab:iqa}
\begin{tabular}{l c>{\columncolor{lightgray}}c c>{\columncolor{lightgray}}c}
\toprule
Metric & SD3.5 & +TACA & FLUX & +TACA \\
\hline
MUSIQ ↑ & 0.7182 & 0.7210 & 0.7186 & 0.7212\\
MANIQA ↑ & 0.4883 & 0.4921 & 0.5149 & 0.5292\\
\bottomrule
\end{tabular}
\vspace{-1.em}
\end{table}

\noindent\textbf{User Study\quad}We invited 50 participants for our user study. From the T2I-Compbench~\cite{huang2023t2icompbench} dataset, we sampled 25 prompts and generated images using FLUX.1 Dev model with and without our TACA method. These images, along with their corresponding text prompts, were presented to the participants. Participants were asked to indicate their preferred image based on three criteria, namely overall visual appeal, attribute (color/shape/texture) quality, and text-image alignment. The results, as summarized in Table~\ref{tab:user_study_results}, demonstrate that a majority of participants favored the images generated by the model incorporating the TACA method. This suggests that our method yields improvements in text alignment and does not ruin image quality.

\begin{table}[h]
    \vspace{-1em}
    \caption{Results of user study.}
    \vspace{-1em}
    \label{tab:user_study_results}
    \centering
    \begin{tabular}{l c>{\columncolor{lightgray}}c}
    \toprule
    {Evaluation Criteria} & {FLUX} & {FLUX + TACA} \\
    \hline
    Overall & 23.58\% &  {76.42\%} \\
    Attribute Quality& 29.25\% &  {70.75\%} \\
    Text-Image Alignment & 17.75\% & {82.25\%} \\
    \bottomrule
    \end{tabular}
    \vspace{-1em}
\end{table}

\subsection{Ablation Study}\label{sec:ablation-study}
\begin{figure*}[h]
    \centering
    \includegraphics[width=0.55\textwidth]{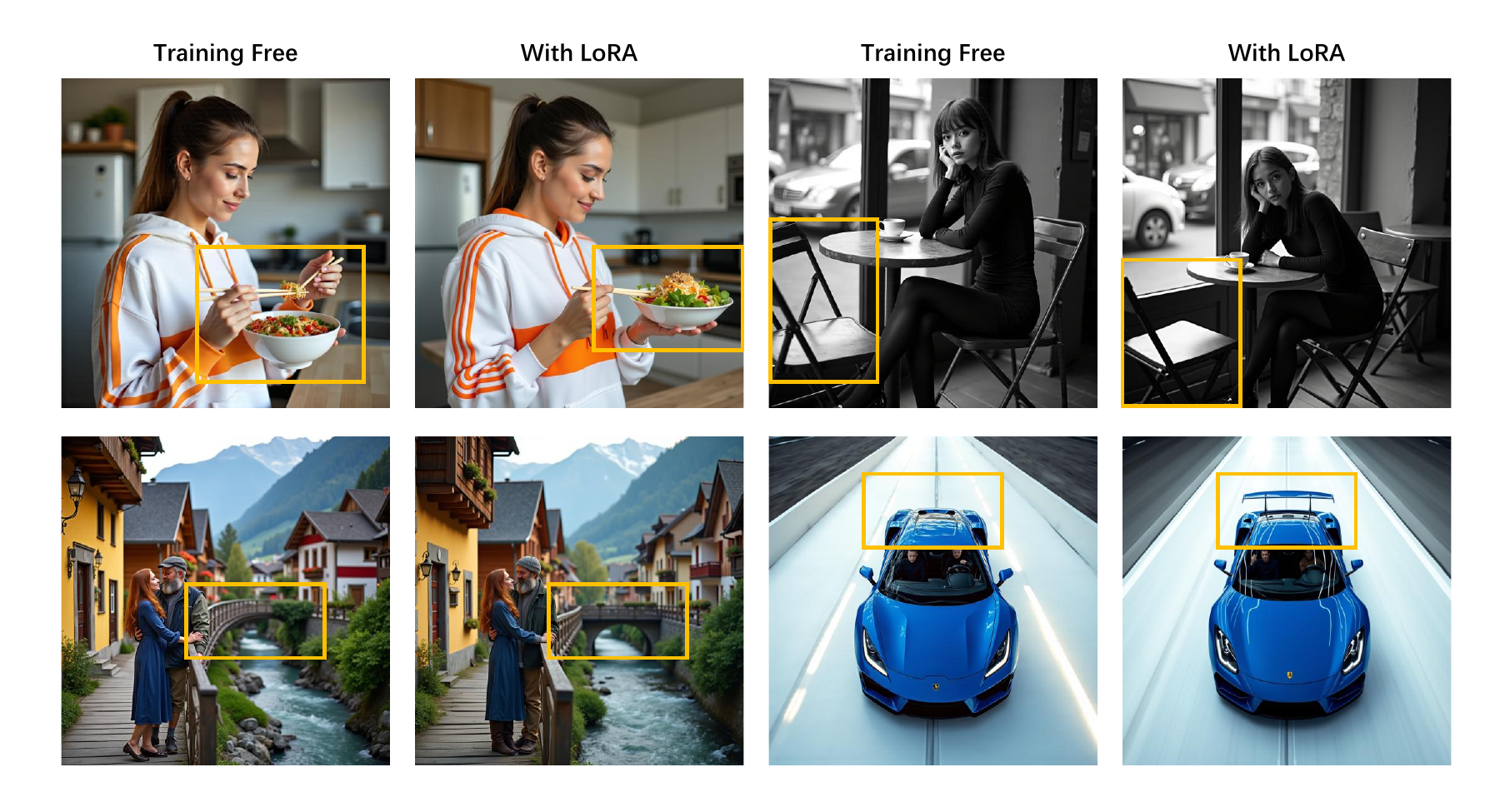}
    \hfill
    \includegraphics[width=0.40\textwidth]{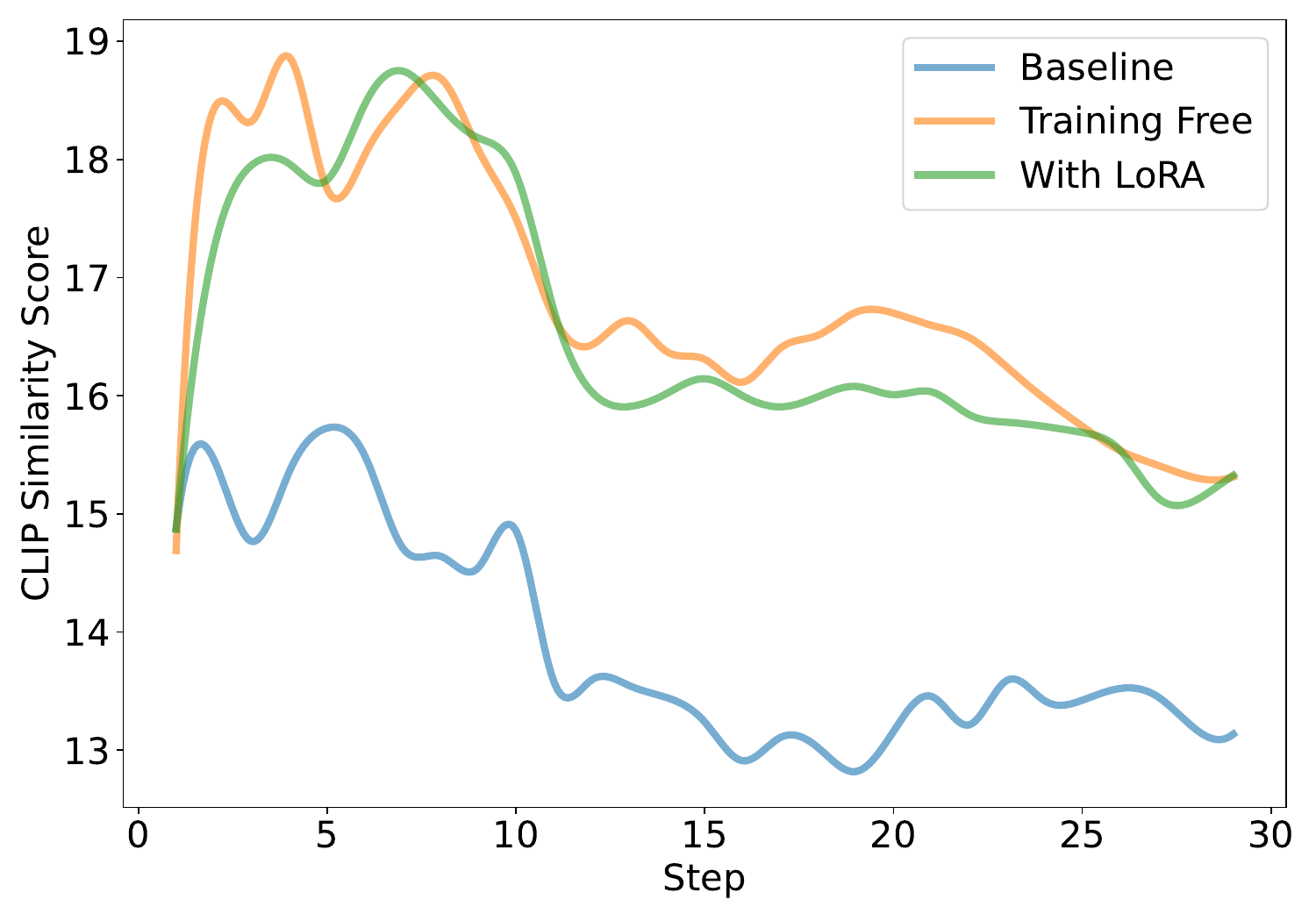}
    \vspace{-1em}
    \caption{The effect of LoRA training. 
    (a) On the left, we show qualitative results comparing training-free image generation to generation with LoRA. The `training-free' examples exhibit artifacts, such as the floating bowl, which are significantly reduced by LoRA training. 
    (b) On the right, we present a quantitative evaluation of CLIP Scores for training-free and LoRA-trained models across the denoising steps, demonstrating that LoRA maintains strong text-image alignment and does not detract from the semantic benefits of our approach.}
    \vspace{-1.5em}
    \label{fig:training-free-vs-lora}
\end{figure*}

\noindent\textbf{Effect of Temperature Scaling Factor ($\gamma_0$)\quad}The temperature scaling factor $\gamma_0$ plays a crucial role in modulating the influence of textual guidance during the denoising process. We explore the impact of different $\gamma_0$ values on compositional generation performance. Table \ref{tab:effect-of-gamma} presents the results.

\begin{table}[h]
\vspace{-0.5em}
\caption{Ablation study on the effect of the temperature scaling factor ($\gamma_0$). We randomly sampled 100 prompts for each attribute from the T2I-CompBench dataset to conduct the evaluation. Here ``LoRA Only'' refers to training a LoRA model solely on the identical dataset using the same hyperparameters without our proposed method. \textbf{Bold} indicates the best score and \underline{underline} indicates the second best score for each attribute.}
\vspace{-0.5em}
\label{tab:effect-of-gamma}
\centering
\small
\begin{tabular}{l cccc}
\toprule
Model           & Color $\uparrow$ & Shape $\uparrow$ & Texture $\uparrow$& Spatial $\uparrow$\\ \hline
FLUX.1-Dev        & 0.798 & 0.591 & 0.755  & 0.193  \\
LoRA Only & \underline{0.805} & 0.592 & 0.759  & 0.187  \\
Ours ($\gamma_0 = 1.1$) & 0.803 & 0.603 & \underline{0.780}  & 0.199  \\
Ours ($\gamma_0 = 1.2$) & \textbf{0.839} & \underline{0.634} & \textbf{0.790}  & \underline{0.207}  \\
Ours ($\gamma_0 = 1.3$) & 0.787 & \textbf{0.650} & 0.766  & \textbf{0.225}  \\ \bottomrule
\end{tabular}\
\vspace{-0.5em}
\end{table}

The results in Table \ref{tab:effect-of-gamma} demonstrate that our proposed method consistently outperforms both the baseline FLUX.1-Dev and the LoRA-only approach across all attributes for reasonable values of $\gamma_0$. We observe improvements in Color, Shape, Texture, and Spatial compositional accuracy. Notably, $\gamma_0=1.2$ yields the best overall balance, achieving the highest scores in Color and Texture, and the second-best in Shape and Spatial. Increasing $\gamma_0$ further to 1.3 leads to slight improvements in Shape and Spatial, but a decline in Color and Texture. This suggests that a moderate increase in textual influence is beneficial, but excessive amplification can negatively impact certain aspects of compositional generation.

To understand the mechanism behind this improvement, we analyze the CLIP similarity between the predicted intermediate latent representations and the text prompt at each denoising step for varying $\gamma_0$. As shown in Figure \ref{fig:training-free-vs-lora} (b), increasing $\gamma_0$ leads to a higher CLIP similarity, particularly in the initial denoising steps. This indicates that TACA effectively enhances the text-image alignment early in the generation process, guiding the model towards generating images that are more consistent with the textual description.

\noindent\textbf{Sensitivity of $\gamma_0$ and $t_{\text{thresh}}$\quad}We further investigate the sensitivity of our method to the choice of $\gamma_0$ and the threshold timestep $t_{\text{thresh}}$ beyond which TACA is applied. Table \ref{tab:gamma-t-sensitivity} presents the average Attribute (Color, Shape, Texture) and Spatial scores for different values of $\gamma_0$ and $t_{\text{thresh}}$ on both FLUX and SD3.5 models.



\begin{table}[H]
\centering
\caption{Sensitivity analysis of $\gamma_0$ and $t_{\text{thresh}}$. The baseline corresponds to the respective models without TACA.}
\vspace{-0.5em}
\resizebox{\linewidth}{!}{  
\begin{tabular}{c cc cc c cc cc}
\toprule
\multirow{2}{*}{$\gamma_0$} & \multicolumn{2}{c}{FLUX} & \multicolumn{2}{c}{SD3.5} & \multirow{2}{*}{$t_{\text{thresh}}$} & \multicolumn{2}{c}{FLUX} & \multicolumn{2}{c}{SD3.5} \\
& attr & spa & attr & spa & & attr & spa & attr & spa \\
\midrule
\rowcolor{lightgray}
\scriptsize{Baseline} &   0.715   & 0.193    &  0.797    &   0.159  & \scriptsize{Baseline} & 0.715   & 0.193    &  0.797    &   0.159    \\
1.15  &   0.740   & 0.197    &   0.810   &   0.181  & 970   &   0.754   &  0.207   &   0.815   &  0.183   \\
1.20  &   0.754   &    0.207 &   0.815   &   0.183  & 950   &   0.757   &  0.218   &    0.811  &  0.172   \\
1.25  &   0.737   &  0.216   &   0.816   &   0.176  & 930   &   0.763   &  0.208   &   0.811   &  0.171   \\
\bottomrule
\end{tabular}
}  
\label{tab:gamma-t-sensitivity}
\end{table}
\vspace{-1.em}

As shown in Table \ref{tab:gamma-t-sensitivity}, the performance exhibits minimal variation across a reasonable range of both $\gamma_0$ (e.g., 1.15 to 1.25) and $t_{\text{thresh}}$ (e.g., 930 to 970). This indicates that our method is not overly sensitive to the precise selection of these parameters, suggesting practical robustness. Furthermore, this robustness is observed across different base models, highlighting the general applicability of the tested parameter ranges.

\noindent\textbf{The Effect of LoRA Training\quad}In the original TACA method, the introduction of factor $\gamma(t)$ induces a shift in the output distribution of each attention layer. These modified outputs are subsequently processed by the feed-forward networks within the transformer blocks. Consequently, the overall output distribution of the diffusion transformer deviates from the distribution inherent in real images, which manifests as visual artifacts like \textit{unsupported floating bowls} and \textit{distorted bridge connections} in Fig~\ref{fig:training-free-vs-lora} (a). To address this issue, we hypothesize that training a LoRA~\cite{Hu2021LoRALA} module can effectively mitigate these artifacts.  The rationale is that by fine-tuning the attention layer weights with a limited number of training samples, the LoRA module enables the modified model to readjust its output distribution to better align with the real image distribution. 

Empirical findings from our experiments demonstrate that the incorporation of LoRA significantly enhances image quality and effectively mitigates these unrealistic artifacts, as evidenced in Fig.~\ref{fig:training-free-vs-lora} (a). Concurrently, we evaluated whether the introduction of LoRA compromises the semantic enhancement facilitated by the temperature coefficient $\gamma(t)$.  The comparative analysis of CLIP Scores for training-free and LoRA configurations for 50 samples, presented in Fig.~\ref{fig:training-free-vs-lora} (b), reveals that LoRA exerts a negligible impact on text-image alignment.
\section{Conclusion and Discussion}
In this paper, we addressed two issues in MM-DiTs that limit text-image alignment in text-to-image generation: suppressed cross-attention due to token imbalance and timestep-insensitive attention weighting. We introduced Temperature-Adjusted Cross-modal Attention (TACA), a simple modification that dynamically balances multimodal interactions using temperature scaling and timestep-dependent adjustment. Combined with LoRA fine-tuning to reduce artifacts, TACA significantly improves text-image alignment on the T2I-CompBench benchmark. Our work demonstrates that strategically reweighting cross-modal interactions leads to more semantically accurate and visually coherent image generation, offering a promising approach for diffusion model research and applications. 

Our work has mainly two limitations: 1) While improvements in text alignment were observed in training-free \textbf{text-to-video} experiments, we encountered a dilution effect when training a LoRA, wherein gains from increasing the temperature factor were diminished. 2) Our method lacks the ability to adaptively select an appropriate scaling factor based on the actual degree of text alignment.
\section{Acknowledgement}
This study is partially supported by the Ministry of Education, Singapore, under its MOE AcRF Tier 2 (MOE-T2EP20221-0012, MOE-T2EP20223-0002), and under the RIE2020 Industry Alignment Fund – Industry Collaboration Projects (IAF-ICP) Funding Initiative, as well as cash and in-kind contribution from the industry partner(s).

{
    \small
    \bibliographystyle{ieeenat_fullname}
    \bibliography{main}
}

\clearpage
\clearpage
\appendix
\setcounter{page}{1}
\maketitlesupplementary

\noindent\textbf{Overview.} In the supplementary material, we provide further details to support our work. Section~\ref{sec:supp-code} elaborates on the implementation of TACA, including code snippets and a speed comparison of different approaches. Section~\ref{sec:supp-abla} presents additional ablation studies focusing on text alignment, examining the effect of CFG guidance scale and the content/length of prompts. Section~\ref{sec:why-lora} explains why we choose LoRA rather than full-parameter finetune. Finally, Section~\ref{sec:supp-examples} showcases more qualitative results with visual comparisons on both short and long prompts using FLUX.1 Dev and SD3.5 Medium.

\section{Code Implementation Details}\label{sec:supp-code}
Given that TACA necessitates modifications to the attention mechanism, and that the functions for computing attention are typically encapsulated within pre-compiled C/C++ binary libraries, directly reimplementing these attention computation functions using PyTorch would result in a significant performance degradation. To minimize the performance impact of modifying the attention mechanism while retaining the convenience of PyTorch, the following two implementation approaches for TACA can be adopted:

\textbf{Flex Attention}
\begin{lstlisting}[caption=PyTorch Flex Attention]
from torch.nn.attention.flex_attention import flex_attention
gamma = 1.2
encoder_size = 512 # T5 encoder seq_len for FLUX

def score_mod(score, batch, head, token_q, token_kv):

    condition = (token_q >= encoder_size) & (token_kv < encoder_size)
    score = torch.where(condition, score * gamma, score)
    return score

hidden_states = flex_attention(query, key, value, score_mod=score_mod)
\end{lstlisting}

\textbf{Selective Attention Recomposition}
\begin{lstlisting}[caption=Selective Attention Recomposition]
gamma = 1.2
encoder_size = 512 # T5 encoder seq_len for FLUX
key_scaled = key.clone()

# Shape of Q, K, V (B, H, N, D)
key_scaled[:, :, :encoder_size, :] *= gamma

# You can also change this into flash attention
hidden_states = F.scaled_dot_product_attention(
    query, key_scaled, value, attn_mask=attention_mask, dropout_p=0.0, is_causal=False
)

hidden_states_orig = F.scaled_dot_product_attention(
    query, key, value, attn_mask=attention_mask, dropout_p=0.0, is_causal=False
)

hidden_states[:, :, :encoder_size, :] = hidden_states_orig[:, :, :encoder_size, :]
\end{lstlisting}

We conducted empirical evaluations of the computational speed of both proposed methods, comparing them against PyTorch's native scaled dot-product attention implementation. All experiments employ a 30-step denoising process to generate $1024 \times 1024$ images via FLUX.1 Dev on a single Nvidia A100 80G GPU. We recorded the performance differential for both a single denoising step and for the complete 30-step denoising process (assuming temperature factor $\gamma$ modification applied only to the initial 10\% of steps). The results of this speed evaluation are presented in Table~\ref{tab:speed_comparison}.

\begin{table}[h]
    \centering
    \begin{tabular}{l|ccc}
        \hline
        \textbf{Method} & \textbf{Single Step} & \textbf{All 30 Steps} & \textbf{Speedup} \\
        \hline
        Baseline       & 0.47 sec                      & 14 sec              & 1.0x                    \\
        Flex            & 2.13 sec                         & 19 sec             & 0.74x                    \\
        Selective &     0.95 sec                     & 16 sec             & 0.88x                   \\
        \hline
    \end{tabular}
    \caption{Speed Comparison of Different Approaches}
    \label{tab:speed_comparison}
\end{table}

\section{Further Ablation Study on Text Alignment}\label{sec:supp-abla}
\begin{table*}[ht]
\centering
\begin{tabular}{ll|cccc}
\hline
Model & Settings & Color $\uparrow$ & Shape $\uparrow$ & Texture $\uparrow$& Spatial $\uparrow$\\
\hline
\multirow{6}{*}{FLUX.1 Dev} & CFG = 3.5 (Default) & 0.798 & 0.591 & 0.755 & 0.193 \\
& CFG = 3.5 + TACA & \textbf{0.839} & \underline{0.634} & \textbf{0.790} & \underline{0.207} \\
\cline{2-6}
& CFG = 5 & 0.787 & 0.553 & 0.756 & 0.175 \\
& CFG = 5 + TACA & \underline{0.835} & \textbf{0.635} & \underline{0.757} & \textbf{0.224} \\
\cline{2-6}
& CFG = 10 & 0.667 & 0.571 & 0.740 & 0.137 \\
& CFG = 10 + TACA & 0.751 & 0.633 & 0.699 & 0.191 \\
\midrule
\multirow{4}{*}{SD3.5-Medium} & CFG = 7 (Default) & 0.812 & 0.730 & 0.850 & 0.159 \\
& CFG = 7 + TACA & \textbf{{0.843}} & \underline{0.737} & \textbf{{0.864}} & 0.183 \\
\cline{2-6}
& CFG = 10 & 0.804 & 0.727 & 0.853 & \underline{0.191} \\
& CFG = 10 + TACA & \underline{0.820} & \textbf{{0.765}} & \underline{0.863} & \textbf{{0.206}} \\
\hline
\end{tabular}
\caption{
Ablation study on the effect of CFG scale with and without TACA (with $\gamma_0 = 1.2$) on FLUX.1 Dev and SD3.5-Medium. We randomly sampled 100 prompts for each attribute from the T2I-CompBench dataset to conduct the evaluation. 
For both models, \textbf{bold} indicates the best score and \underline{underline} indicates the second-best score for each attribute.
}
\label{tab:effect-of-cfg}
\end{table*}
\subsection{The scale of CFG guidance}
To investigate the text alignment improvements offered by our TACA method in comparison to increasing the CFG guidance scale (commonly employed in text-to-image models to enhance alignment, often at the cost of image quality), we conducted a series of ablation studies. These experiments aimed to determine whether TACA maintains its efficacy across varying CFG guidance scales and across different models. The results, presented in Table~\ref{tab:effect-of-cfg}, reveal the effects of different CFG scales and the impact of TACA on both FLUX.1 Dev and SD3.5-Medium.

For FLUX.1 Dev, the default guidance scale of 3.5 appears to be a ``sweet spot'': further increases in CFG intensity beyond this point yield minimal gains in text alignment, and, notably, performance across several metrics degrades significantly. Concurrently, our TACA method demonstrated effectiveness across diverse guidance scales, suggesting its general applicability.

For SD3.5-Medium, increasing the CFG scale also enhances text-image alignment but tends to degrade visual fidelity, resulting in reduced metrics (e.g., Color score drops at CFG=10 compared to CFG=7). Our TACA method, however, directly reinforces the dependence of image tokens on textual tokens, improving alignment without such adverse effects. TACA consistently improves results across different CFG scales on SD3.5-Medium, showing both generalization and complementarity.

Overall, the combined results across both models indicate that while increasing CFG can improve alignment to some extent, it often comes at the cost of overall performance. TACA, on the other hand, offers a more targeted and effective approach to enhancing text-image alignment, being beneficial and complementary across different CFG scales and diffusion models.

\subsection{The content of the prompt}
We have identified several prevalent issues regarding text alignment in state-of-the-art text-to-image models. Our TACA can mitigate these issues to a certain extent.

\begin{itemize}
    \item Difficulty in handling unrealistic scenarios, such as \textit{``a blue sun and a yellow sea''}.
    \item Difficulty in handling spatial relationships, such as with the prompt \textit{``a squirrel to the left of the man''}. Models frequently interpret the left side of the image as the left side specified in the text, rather than the left side relative to the man's frame of reference within the image.
    \item Difficulty in handling specific numerical quantities. For instance, when prompted for four vases, the model may generate images containing five or three vases.
\end{itemize}

\subsection{The length of the prompt}
We also observe that models are more prone to omitting details from longer prompts, particularly when the prompt's token count exceeds the maximum token limit supported by the CLIP text encoder.

\begin{figure*}
    \centering
    \includegraphics[width=0.85\linewidth]{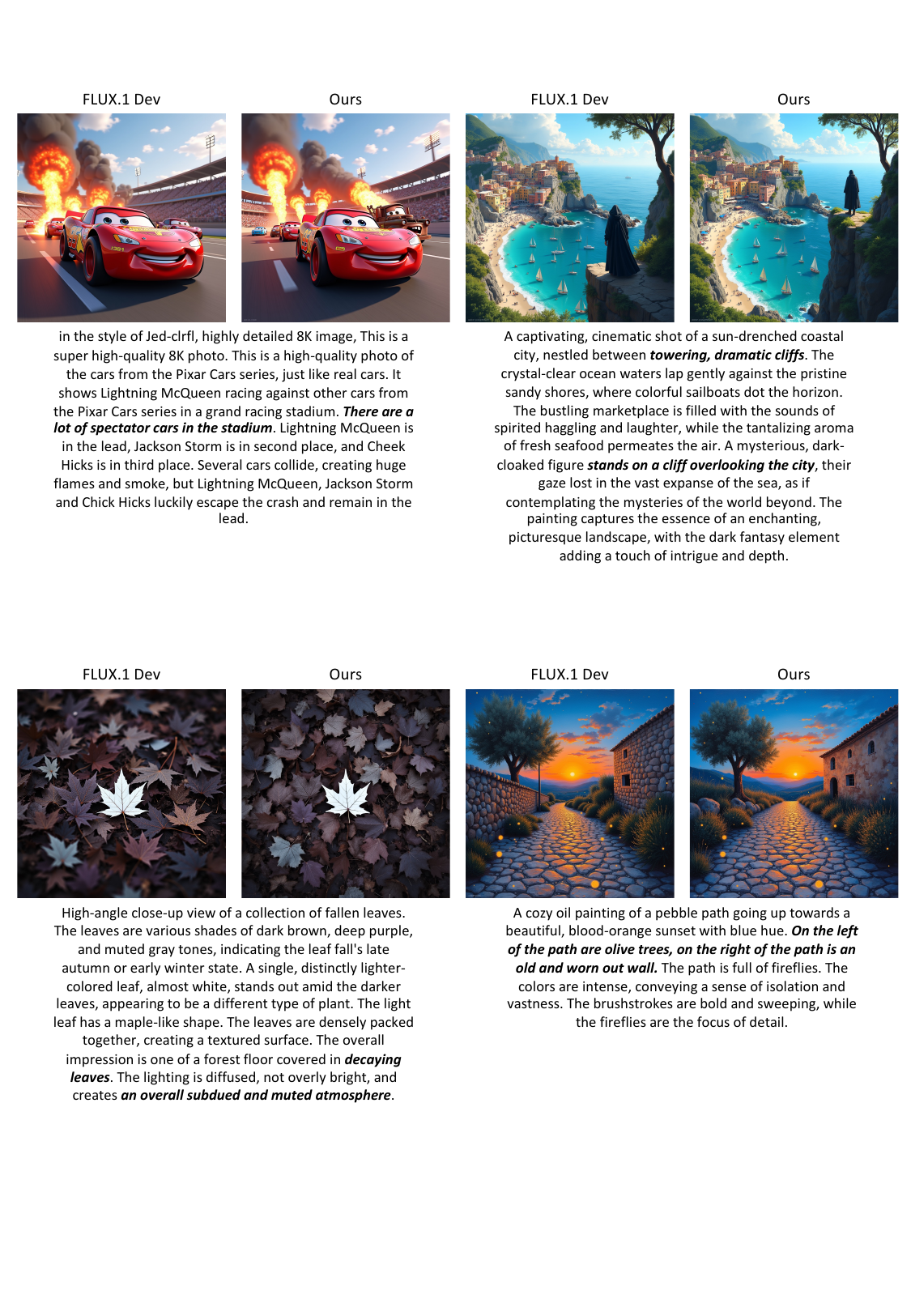}
    \caption{Visual comparisons on text-image alignment (FLUX.1 Dev, long prompts)}
    \label{fig:sup-7}
\end{figure*}

\begin{figure*}
    \centering
    \includegraphics[width=0.85\linewidth]{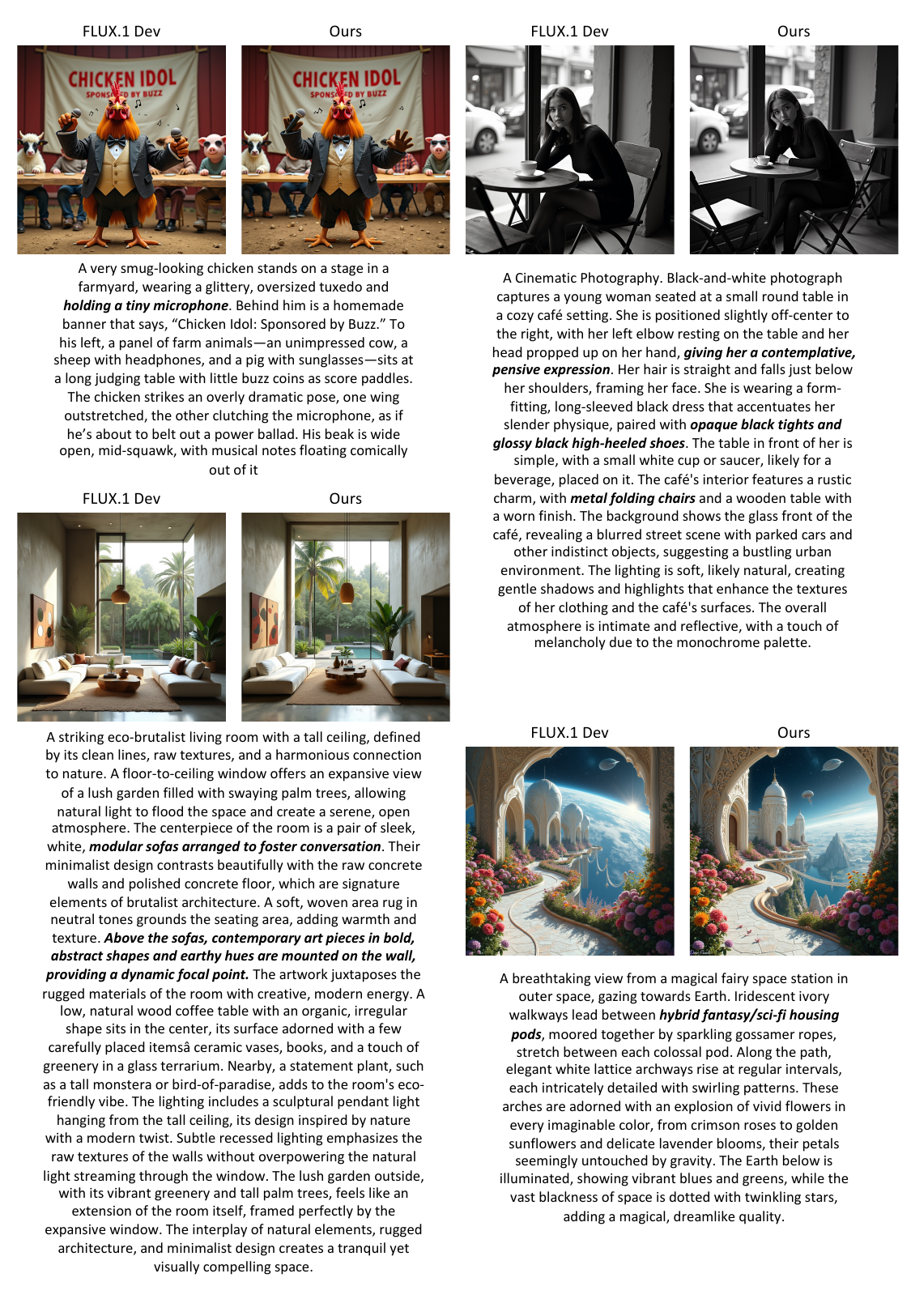}
    \caption{Visual comparisons on text-image alignment (FLUX.1 Dev, long prompts)}
    \label{fig:sup-8}
\end{figure*}

\begin{figure*}
    \centering
    \includegraphics[width=0.85\linewidth]{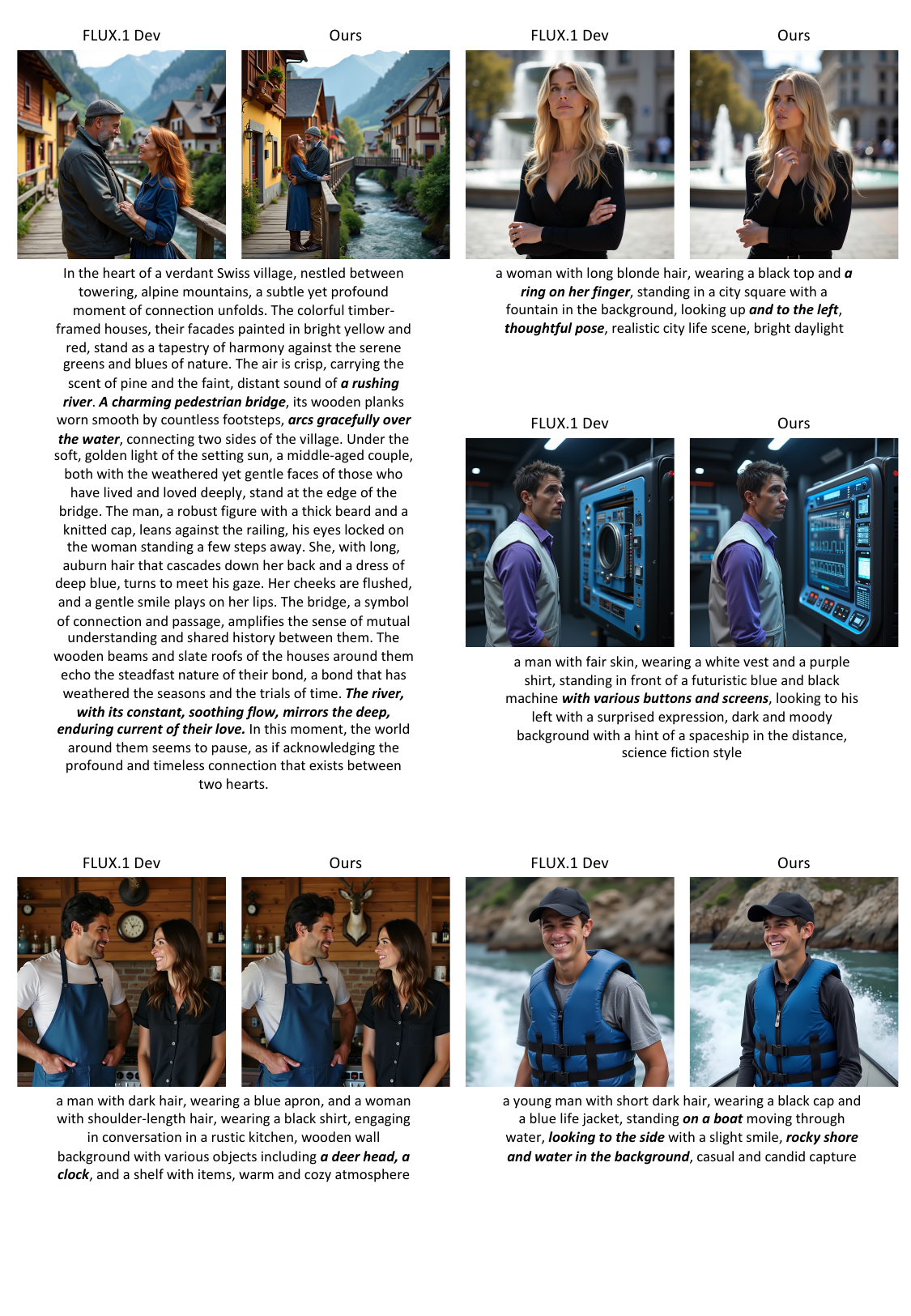}
    \caption{Visual comparisons on text-image alignment (FLUX.1 Dev, long prompts)}
    \label{fig:sup-9}
\end{figure*}

Our proposed TACA method demonstrates comparatively more widespread effectiveness for mitigating the attribute missing issues often found in longer prompt, rather than the shorter ones. Currently, a mature benchmark for evaluating the text-image alignment capabilities of text-to-image models with long prompts is lacking, despite the practical prevalence of longer prompts in real-world applications. Therefore, we have manually curated a set of authentic, long prompts from the internet to assess our method's performance, and the corresponding results are presented in Fig.~\ref{fig:sup-7,fig:sup-8,fig:sup-9}.

\section{Full parameter fine-tuning vs LoRA}\label{sec:why-lora}
In addition to LoRA training, we also experimented with full parameter fine-tuning as an alternative approach. However, we found that this method required significantly more computational resources and storage, especially for large models like FLUX.1 Dev. Moreover, our experiments revealed that full parameter fine-tuning is highly sensitive to learning rate settings. If the learning rate is set too high, the generated images tend to appear blurry or overly stylized, resembling oil paintings. On the other hand, if the learning rate is too low, the model struggles to learn the original data distribution effectively. These challenges, combined with the lack of superior artifact reduction compared to LoRA, led us to conclude that LoRA training is a more robust, efficient, and practical solution.
\section{More Qualitative Results}\label{sec:supp-examples}

\begin{figure*}
    \centering
    \includegraphics[width=0.85\linewidth]{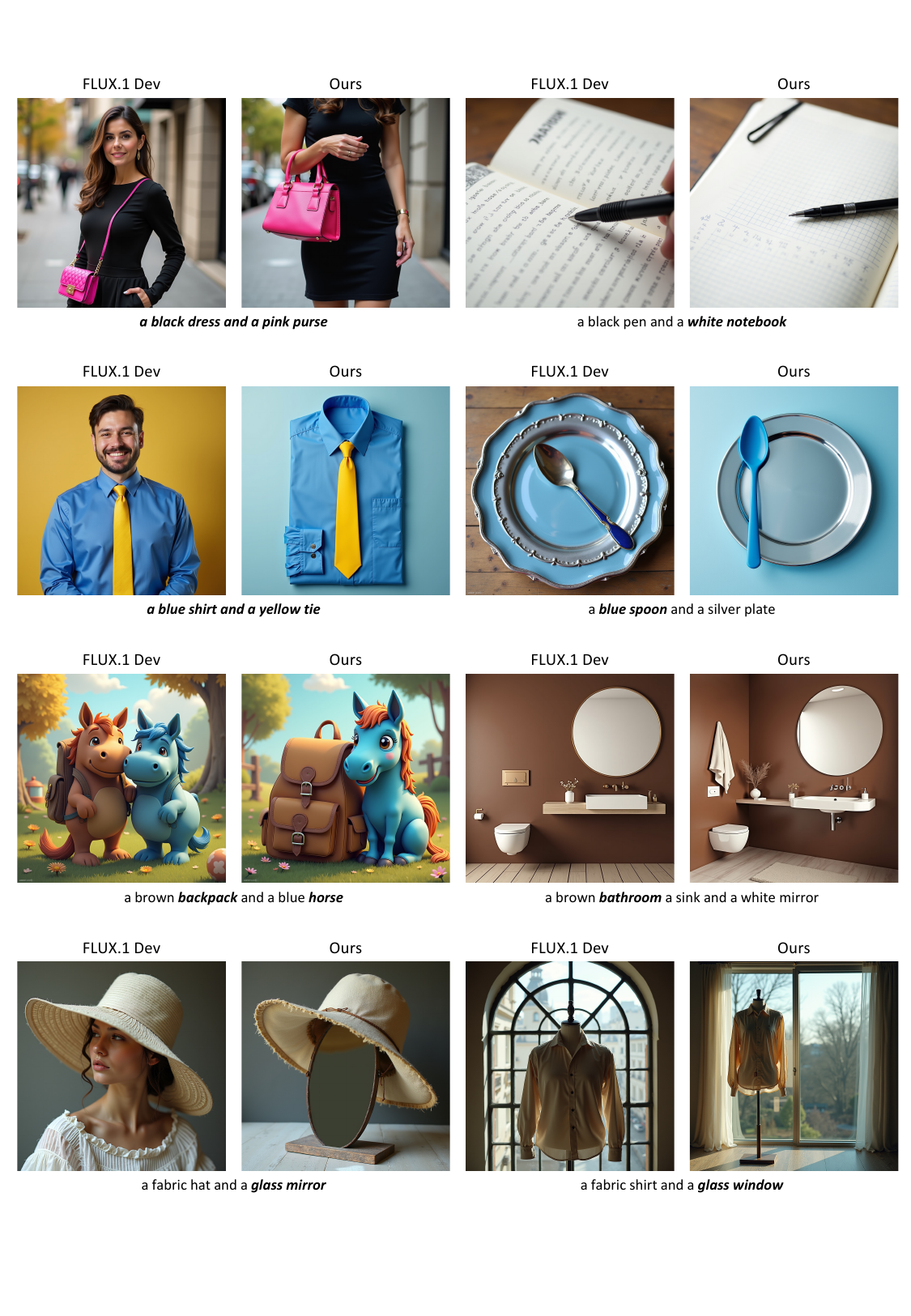}
    \caption{Visual comparisons on text-image alignment (FLUX.1 Dev, short prompts)}
    \label{fig:sup-1}
\end{figure*}

\begin{figure*}
    \centering
    \includegraphics[width=0.85\linewidth]{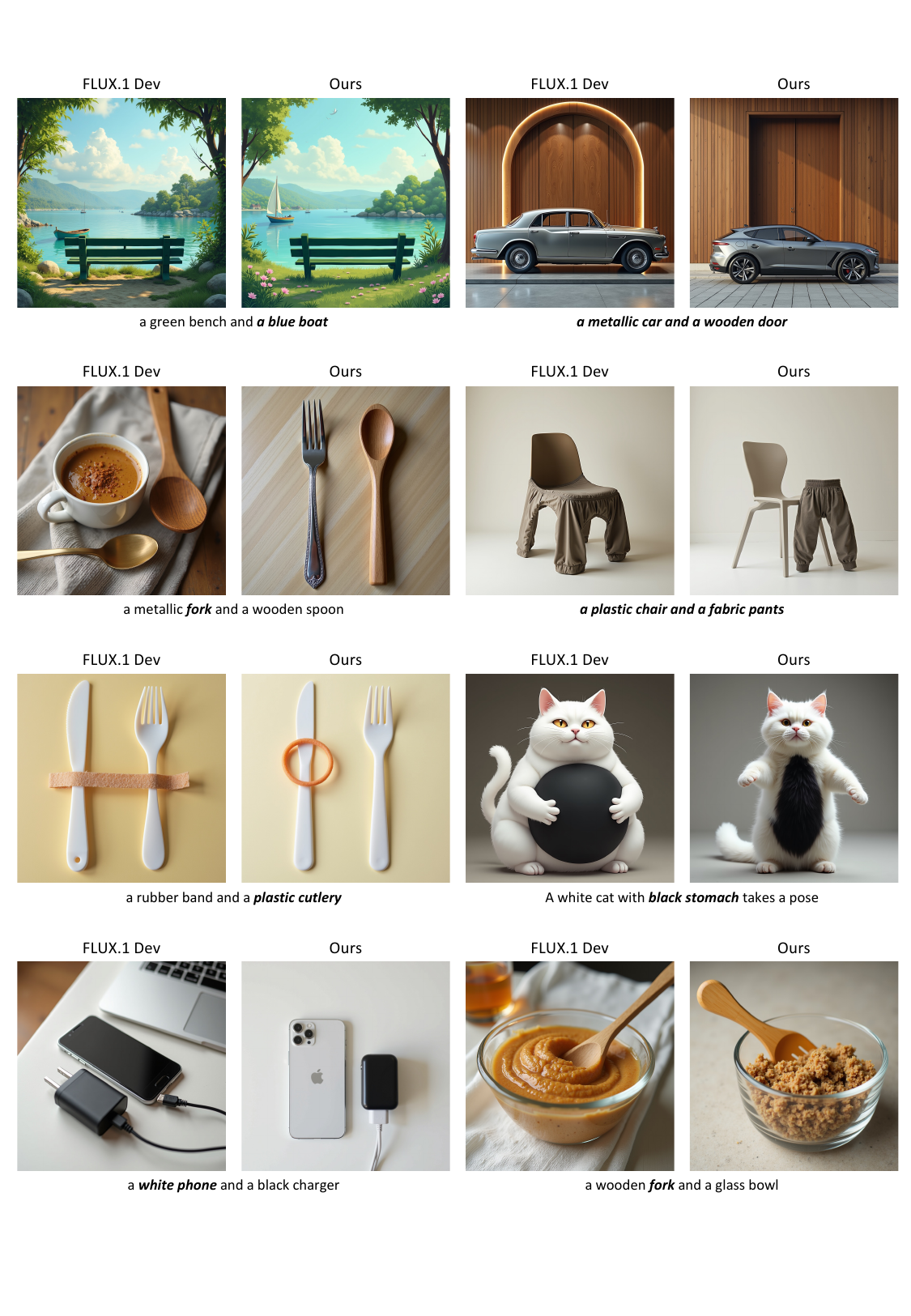}
    \caption{Visual comparisons on text-image alignment (FLUX.1 Dev, short prompts)}
    \label{fig:sup-2}
\end{figure*}

\begin{figure*}
    \centering
    \includegraphics[width=0.85\linewidth]{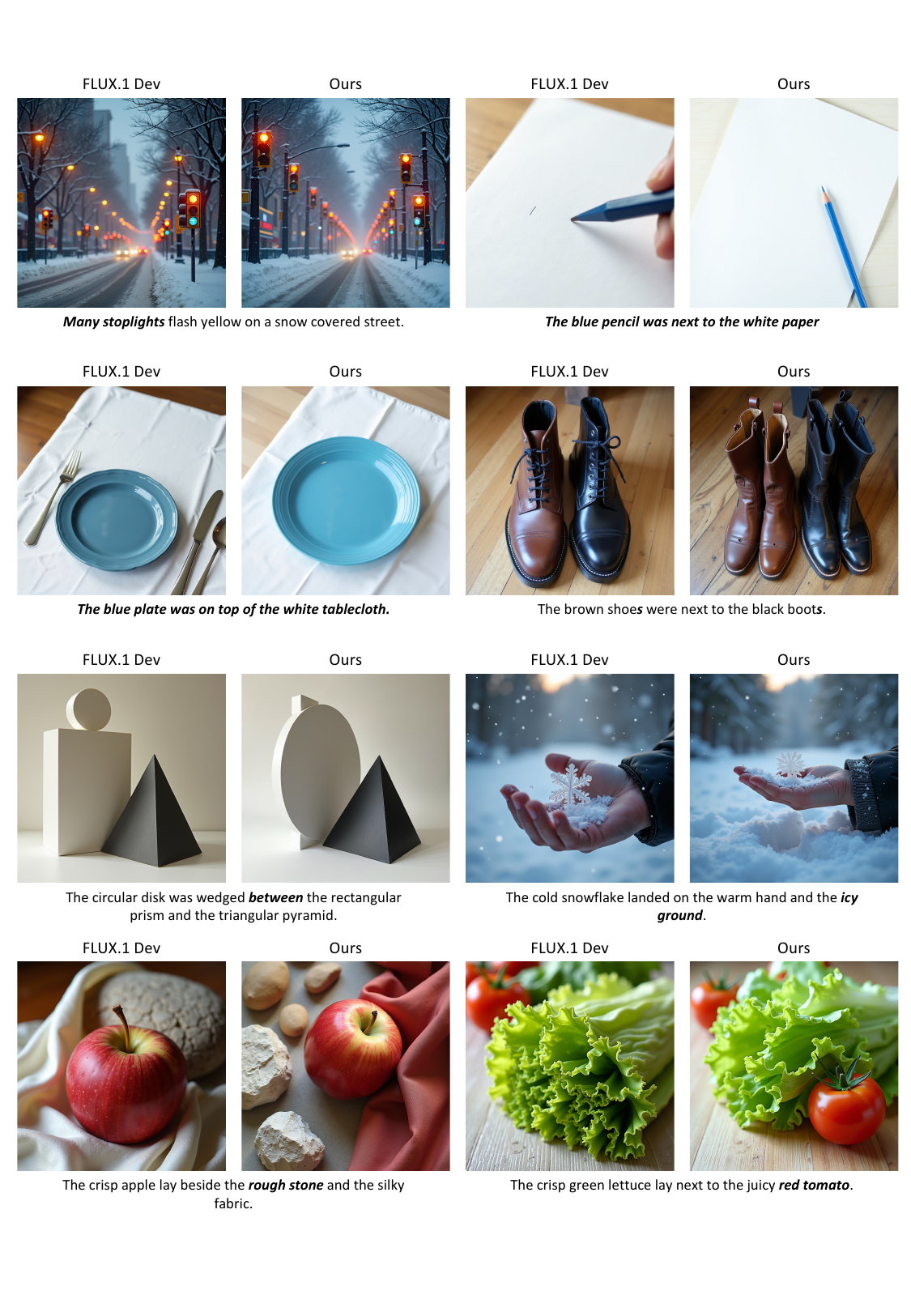}
    \caption{Visual comparisons on text-image alignment (FLUX.1 Dev, short prompts)}
    \label{fig:sup-3}
\end{figure*}

\begin{figure*}
    \centering
    \includegraphics[width=0.85\linewidth]{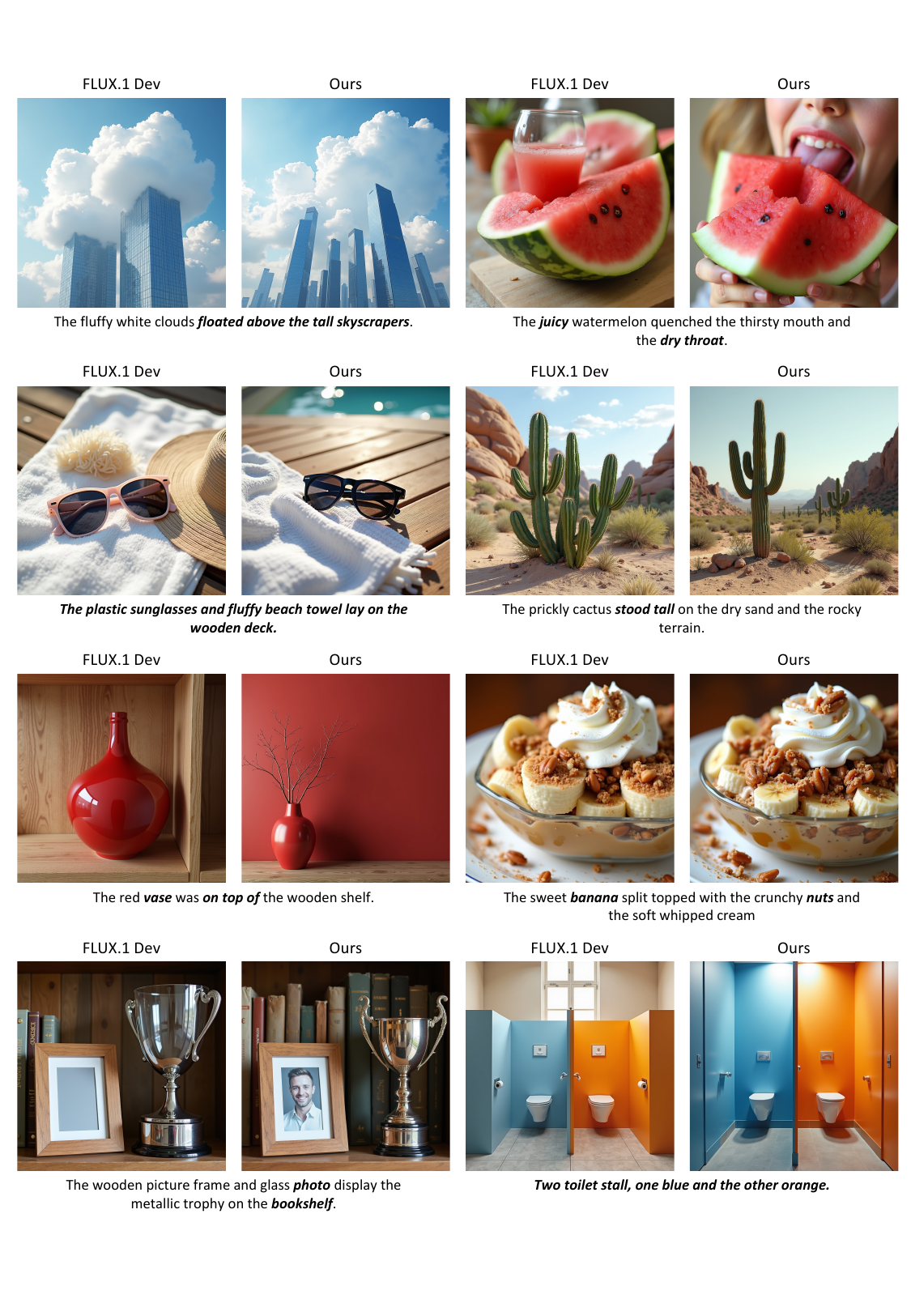}
    \caption{Visual comparisons on text-image alignment (FLUX.1 Dev, short prompts)}
    \label{fig:sup-4}
\end{figure*}

\begin{figure*}
    \centering
    \includegraphics[width=0.85\linewidth]{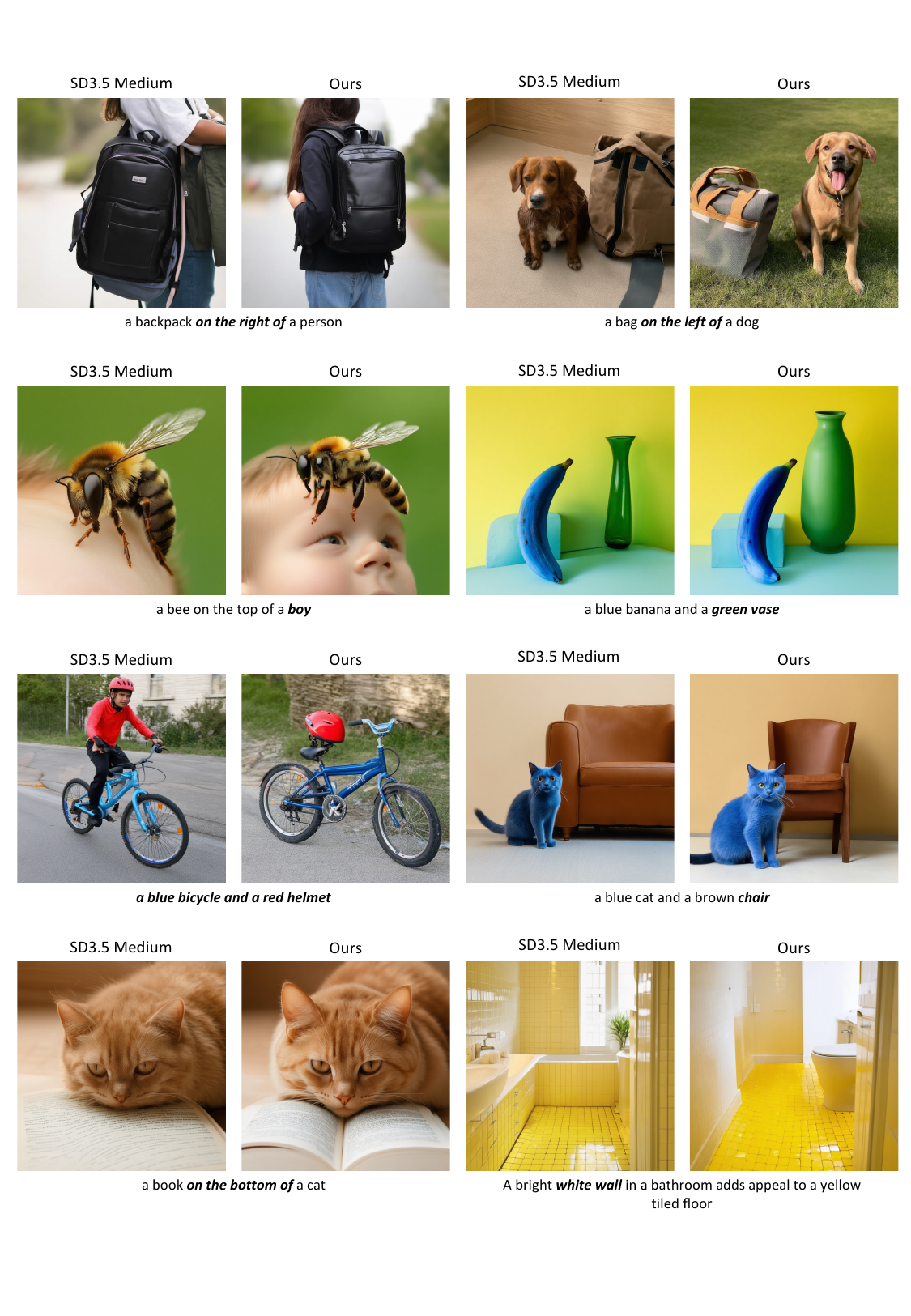}
    \caption{Visual comparisons on text-image alignment (SD3.5 Medium, short prompts)}
    \label{fig:sup-5}
\end{figure*}

\begin{figure*}
    \centering
    \includegraphics[width=0.85\linewidth]{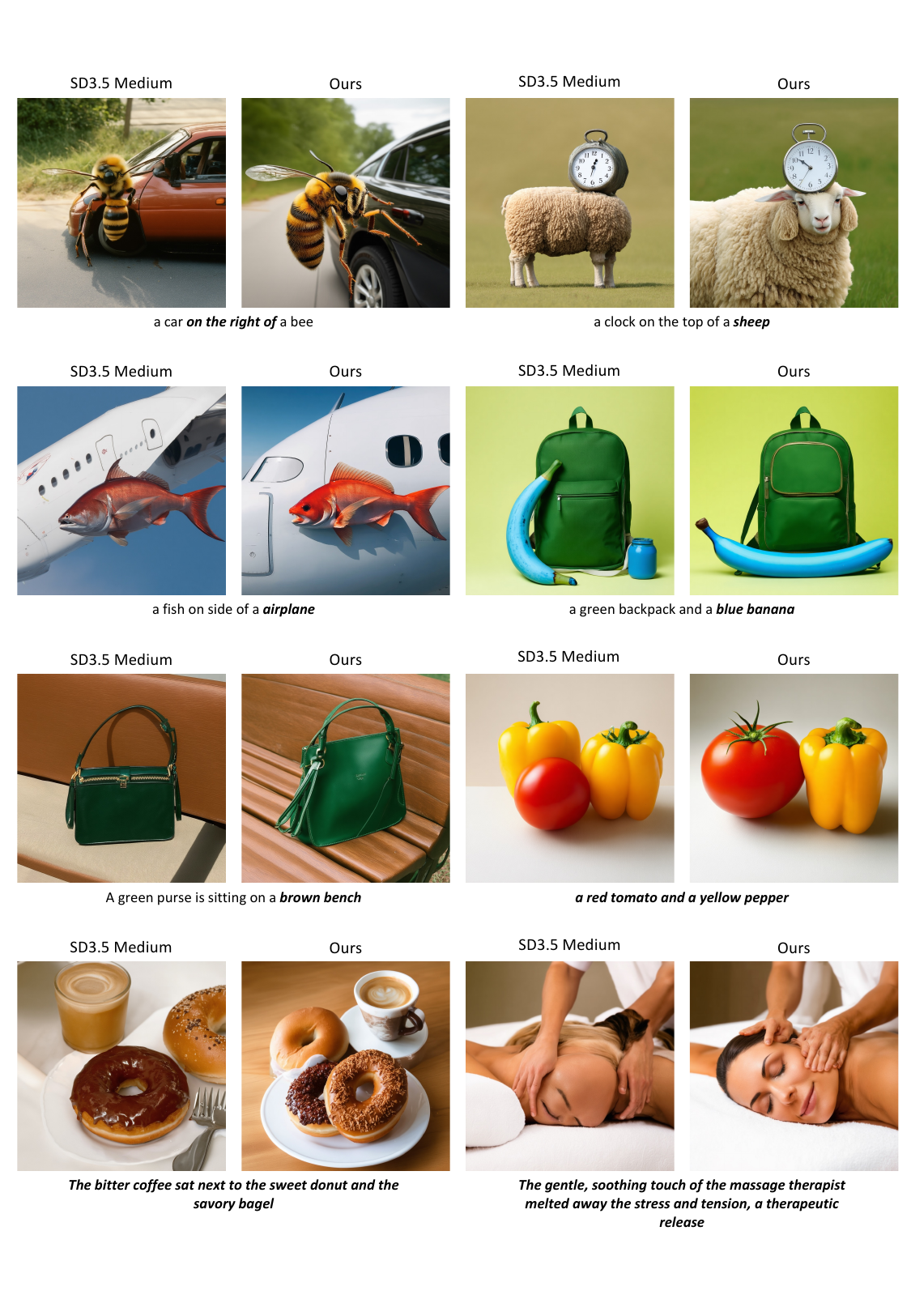}
    \caption{Visual comparisons on text-image alignment (SD3.5 Medium, short prompts)}
    \label{fig:sup-6}
\end{figure*}

\end{document}